%% file: main.tex
\pdfoutput=1
\PassOptionsToPackage{prologue,dvipsnames}{xcolor}  

\documentclass[11pt]{article}

\usepackage[final]{acl}

\usepackage{times}
\usepackage{latexsym}
\usepackage{pgfplots}

\pgfplotsset{compat=1.18}
\usepackage[T1]{fontenc}

\usepackage[utf8]{inputenc}

\usepackage{microtype}
\usepackage{bm}
\usepackage{inconsolata}

\usepackage{graphicx}
\usepackage{subcaption}
\usepackage{amsmath}

\usepackage{adjustbox} 
\usepackage{placeins,booktabs,multirow}
\usepackage[inkscapelatex=false]{svg}

\usepackage{amsmath}
\usepackage{amssymb}

\usepackage[dvipsnames]{xcolor}
\usepackage{relsize}

\expandafter\def\expandafter\normalsize\expandafter{%
    \normalsize%
    \setlength\abovedisplayskip{5pt}%
    \setlength\belowdisplayskip{5pt}%
    \setlength\abovedisplayshortskip{-12pt}%
    \setlength\belowdisplayshortskip{20pt}%
}

\title{Compound AI Systems Optimization:\\ A Survey of Methods, Challenges, and Future Directions}

\author{Yu-Ang Lee\thanks{Equal contribution}\quad Guan-Ting Yi\footnotemark[1]\quad Mei-Yi Liu\\\textbf{Jui-Chao Lu\quad Guan-Bo Yang\quad Yun-Nung Chen}\\
  National Taiwan University, Taipei, Taiwan\\
  \texttt{\{r12946015,r13922053,r10228031,b09901142,r13922083\}@ntu.edu.tw}\\
  \texttt{y.v.chen@ieee.org}
  }

\begin{document}
\maketitle
\begin{abstract}
Recent advancements in large language models (LLMs) and AI systems have led to a paradigm shift in the design and optimization of complex AI workflows. By integrating multiple components, compound AI systems have become increasingly adept at performing sophisticated tasks. However, as these systems grow in complexity, new challenges arise in optimizing not only individual components but also their interactions. While traditional optimization methods such as supervised fine-tuning (SFT) and reinforcement learning (RL) remain foundational, the rise of natural language feedback introduces promising new approaches, especially for optimizing non-differentiable systems.
This paper provides a systematic review of recent progress in optimizing compound AI systems, encompassing both numerical and language-based techniques. We formalize the notion of compound AI system optimization, classify existing methods along several key dimensions, and highlight open research challenges and future directions in this rapidly evolving field.\footnote{\url{https://github.com/MiuLab/AISysOpt-Survey}}
\end{abstract}

\input{text/intro.tex}
\input{text/background.tex}
\input{text/ai_system.tex}
\input{text/discussions.tex}

\input{text/conclusion.tex}

\FloatBarrier 

\section*{Limitations}
We acknowledge several limitations of this survey.
First, due to the lack of a universally accepted definition of ``compound AI systems,'' we also include works that self-identify as optimizing multi-agent systems (MAS) or LM programs, without systematically analyzing their conceptual overlaps and distinctions.

 Second, we focus exclusively on methods that explicitly optimize systems of multiple nodes, thereby excluding traditional prompt optimization techniques for standalone LLMs and contributions not framed as system optimization. 
 Despite our efforts to draw a clear boundary, some relevant papers may have been inadvertently omitted; moreover, as this field is actively evolving during the preparation of this manuscript, studies published in the last two months may be only partially covered. To address this, we maintain an open‐source repository 
 where readers can submit works we may have missed. 
 
 Third, due to page limits, we highlight only the core motivation and algorithmic design of each method and omit details such as experimental setups and results. Readers are encouraged to consult the original papers and code repositories for full technical details once they have gained an overview from our survey.

 \section*{Acknowledgements}
We thank the reviewers for their insightful comments.
This work was financially supported by the National Science and Technology Council (NSTC) in Taiwan, under Grants 111-2222-E-002-013-MY3 and 112-2223-E002-012-MY5, and the Center of Data Intelligence: Technologies, Applications, and Systems, National Taiwan University under the Grant 114L900901, from the Featured Areas Research Center Program within the framework of the Higher Education Sprout Project by the Ministry of Education (MOE) of Taiwan.


\bibliography{custom}

\clearpage
\appendix 
\input{appendix/A1}
\input{appendix/A2}

\end{document}

%% file: text/intro.tex
\section{Introduction}
The community has witnessed a new generation of AI systems centered on large language models (LLMs), incorporating several sophisticated components such as simulators, code interpreters, web search tools, and retrieval-augmented generation (RAG) modules. These systems have shown remarkable capabilities across domains and typically outperform standalone LLMs. For instance, LLMs communicating with symbolic solvers can tackle Olympiad-level math problems~\cite{trinh2024solving}, integrate with search engines and code interpreters to match human programmers’ performance~\cite{li2022competition,yang2024swe}, and when coupled with 
knowledge graphs, drive
biological materials discovery~\cite{ghafarollahi2024sciagents}.

\begin{figure}[t]
\centering
\includegraphics[width=\linewidth]{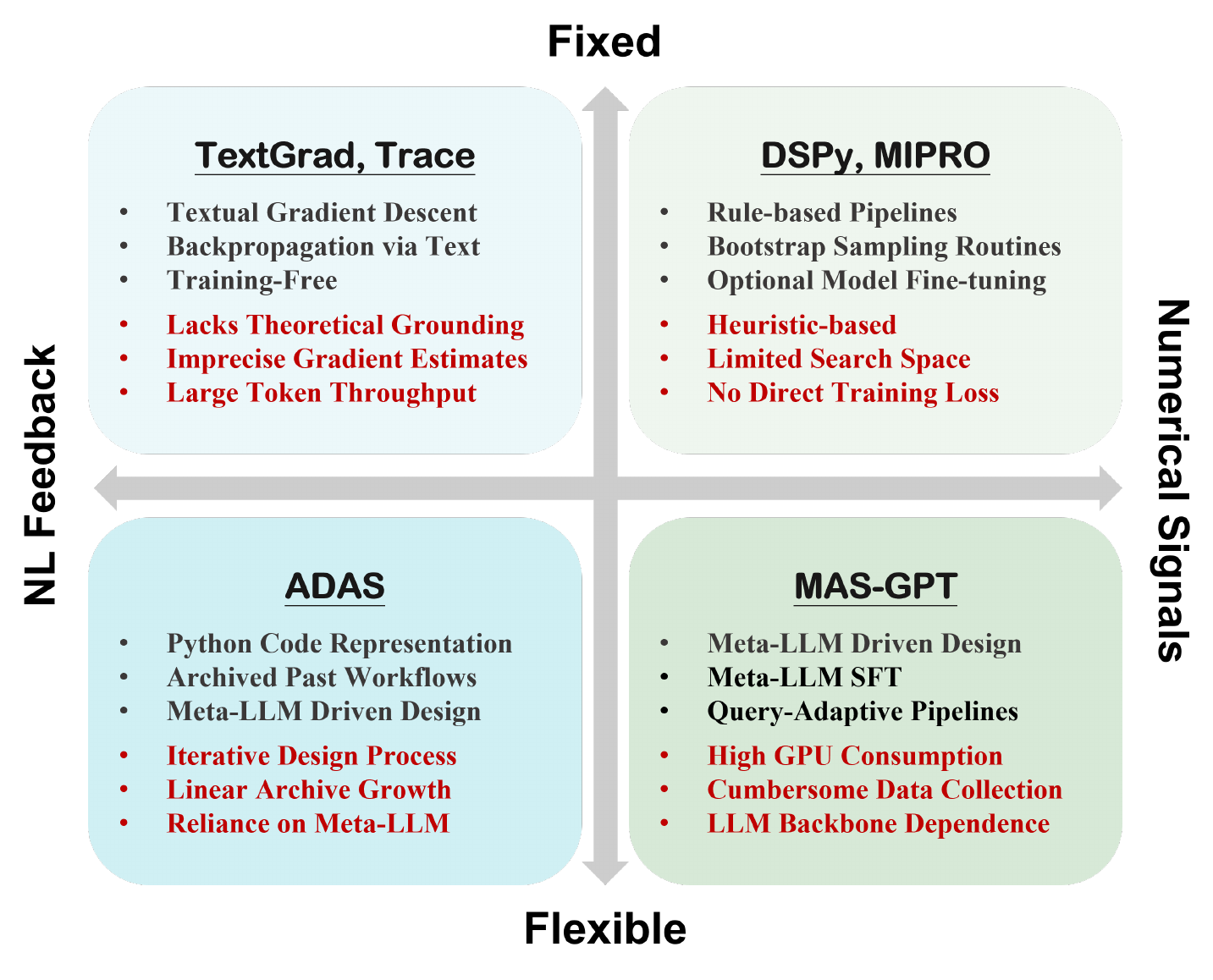}
  \caption{\textbf{The proposed 2$\times$2 taxonomy} spans Structural Flexibility (y-axis) and Learning Signals (x-axis). Representative methods for each quadrant along with their key designs and potential \color{BrickRed}drawbacks\color{black}.}
  \label{fig:1}
\end{figure}

\begin{figure*}[t]
\centering
\includegraphics[width=0.95\textwidth, height=0.34\textwidth]{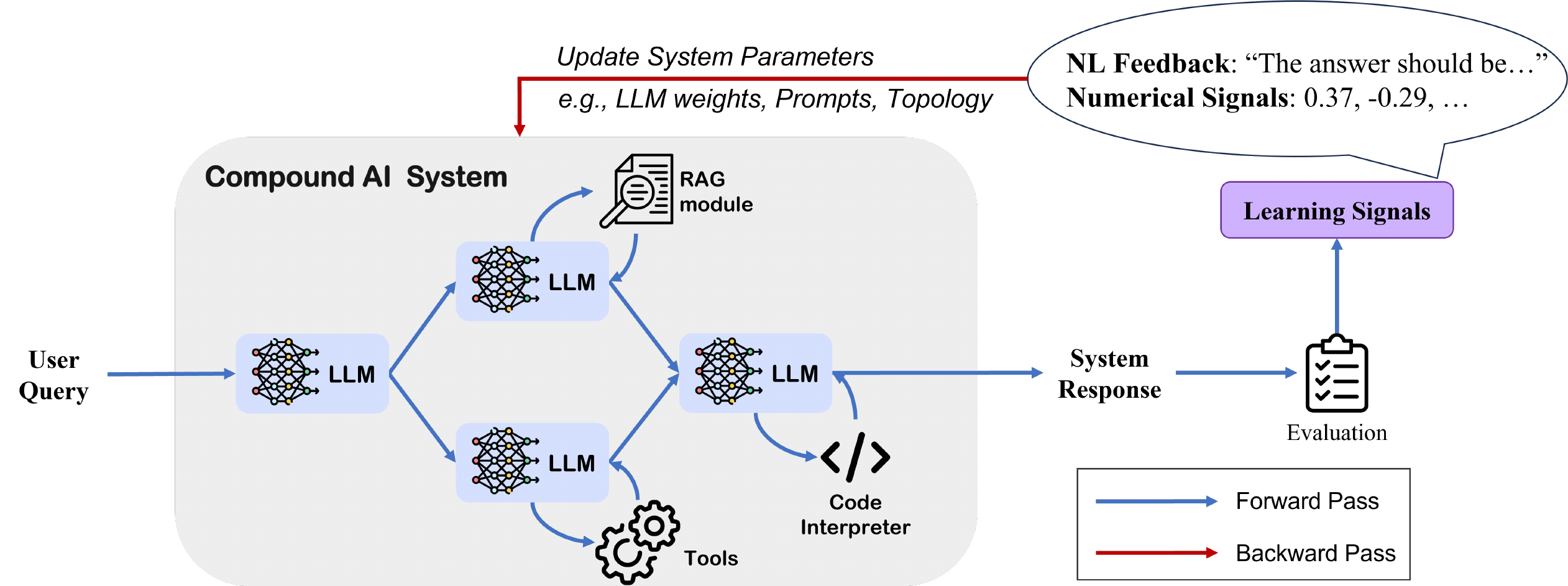}
  \caption{\textbf{Example of a compound AI system and its optimization.}  
Centered on LLMs and coupled with multiple interacting modules, the system handles complex user queries. Automated optimization strategies leverage two types of learning signals, i.e., natural language feedback and numerical signals (defined in Sec.~\ref{sec:four_dim}), to backpropagate errors and guide system updates toward improved performance.}
  \label{fig:sys_demo}
\end{figure*}

However, even with mature toolkits that streamline the design process of compound AI systems, such as, LangChain~\cite{Chase_LangChain_2022} and LlamaIndex~\cite{Liu_LlamaIndex_2022}, substantial human intervention remains essential to tailor these systems toward targeted downstream applications~\cite{xia2024agentless, zhang2024chain}, often involving trial-and-error tuning prompt templates and system pipelines based on heuristics. This limitation has motivated recent efforts to develop principled, automated methods for end-to-end AI system optimization. Yet, the operational schemes of these approaches diverge significantly depending on whether they permit modifications to the system topology and how they transmit learning signals. Moreover, the field still lacks standardized terminology or a cohesive conceptual framework, making some articles difficult for newcomers to navigate~\cite{cheng2024_trace, khattab2023_dspy}. Despite existing surveys~\cite{lin2024_llm_opt, liu2025advancesagents} have provided helpful frameworks, they focus on optimization driven by natural language, overlook critical schemes that permit updates to system topology and do not cover the most recent advances.

To address these gaps, this study introduces four key dimensions for examining existing methods and, drawing on two of these dimensions, constructs a 2$\times$2 taxonomy (Fig.~\ref{fig:1}) that holistically covers 26 representative works. Setting a reasonable scope for our survey, we excluded related but not directly relevant work. These include 
prompt optimization techniques for a single LLM
\cite{APE, APO, OPRO, das2024_greater, EvoPropmt}, 
systems built for task-specific applications without optimization~\cite{zhuge2023mindstorms, hong2023metagpt}, and papers that do not explicitly frame their problem as system optimization~\cite{zhou2022least, sordoni2023joint}. The proposed taxonomy is inherently adaptable to future research, and we hope it will enable developers of compound AI systems to gain a high-level overview before delving into technical details, while allowing researchers proposing novel algorithms to better situate and compare their contributions.

The remainder of this paper is organized as follows: Sec.~\ref{sec:back} provides the necessary background and formalization for compound AI systems and their optimization. Sec.~\ref{sec:four_dim} discusses the four key dimensions, followed by Sec.~\ref{sec:review}, which presents a detailed review of surveyed papers based on the proposed 2$\times$2 taxonomy. Sec.~\ref{sec:discussions} identifies the main challenges facing current methods and their corresponding future directions, and Sec.~\ref{sec:conclusion} concludes with a summary of our contributions.

%% file: text/background.tex
\section{Background and Preliminaries}\label{sec:back}
\subsection{Compound AI Systems}
In contrast to single AI models that function as statistical models (e.g., the Transformer~\cite{vaswani2017attention} for next-token prediction), compound AI systems are defined as systems that tackle AI tasks using multiple interacting components~\cite{compound-ai-blog} (see Fig.~\ref{fig:sys_demo}).
The term compound AI system somehow overlaps with related concepts and is often used interchangeably in the field. These include multi-agent systems (MAS)~\cite{zhou2025_multi, wang2025_scoreflow}, language model pipelines~\cite{khattab2023_dspy, soylu2024fine}, and language model programs~\cite{opsahl2024optimizing}. Hence, in this survey, we include any method that aims to optimize AI systems composed of multiple components, and adopt “compound AI system” as a unifying label.

Although end-to-end optimization of single models such as neural networks implemented in PyTorch~\cite{paszke2019pytorch} is straightforward thanks to gradient-based backpropagation~\cite{rumelhart1986learning} on their fully differentiable layer connections, compound AI systems are built from non-differentiable components and thus require novel optimization methods. Representative examples include heuristic bootstrap-based methods~\cite{khattab2023_dspy} applied to find optimal in-context examples in LLM prompts, as well as approaches leveraging an auxiliary LLM to provide textual feedback on prompt updates~\cite{Mert2024_textgrad} or propose improved system topologies~\cite{hu2024_ADAS}.

\subsection{Formal Definitions}\label{sec:formal_def}
Due to inconsistent mathematical descriptions across the surveyed papers, we develop a unified graph-based formalization as well as notations for compound AI systems and their optimization\footnote{If our notations conflict with that of any original work, our definitions take precedence.}.

Specifically, we use $\Phi=(G, \mathcal{F})$ to denote a compound AI system that inputs a user query $q \in \mathcal{Q}$ and outputs a response (answer) $a\in \mathcal{A}$. In particular, $G = (V, E)$ is a directed graph and $\mathcal{F} = \{f_i\}_{i=1}^{|V|}$ is a set of operations (e.g., LLM forward pass, RAG step, external tool calling). Each $f_i$ is attached to node $v_i$, producing 
\begin{equation}
    Y_i = f_i(X_i;\,\Theta_{i}),    
\end{equation}
where $X_i$ is the input to $v_i$, $Y_i$ its output, and $\Theta_i$ its parameters. The edge matrix $E=[c_{ij}]$ comprises Boolean functions $c_{ij}\!:\Omega \to \{0,1\}$ that determine whether the potential edge from $v_i$ to $v_j$ is active. This means given the contextual state $\tau \in \Omega$, the edge $(v_i \!\to\! v_j)$ is active if and only if $c_{ij}(\tau)=1$. Consequently, even though $G$ and $\mathcal{F}$ are fixed after optimization, the $\Phi$’s effective topology varies with $\tau$, reflecting dependencies on both the input query $q$ and intermediate outputs of the system, as visualized in Appendix Fig.~\ref{fig:ai_system}.

In more detail, the node input is set by 
\begin{equation}
   X_i \;\leftarrow\; \mathop{\oplus}\limits_{j:\,c_{ji}(\tau)=1} Y_j,
\end{equation} 
where $\oplus$ denotes concatenation of the outputs $Y_j$ from all nodes $v_j$ whose edge $v_j \to v_i$ is active. The node parameter $\Theta_i$ decomposes as $\Theta_i = (\theta_{i,N}, \theta_{i,T})$, where $\theta_{i,N}$ are numerical parameters (e.g., LLM weights, temperature) and $\theta_{i,T}$ are textual parameters (e.g., LLM prompts). Due to space constraints, descriptions of the input node $v_{\mathrm{in}}$, the output node $v_{\mathrm{out}}$, and our formalism’s ability to either support or prohibit cyclic structures are deferred to Appendix~Sec.~\ref{sec:more_math}. 

Given a training set $\mathcal{D} = \{(q_i, m_i)\}_{i=1}^N$, where each query $q_i$ is associated with optional metadata $m_i \in \mathcal{M}$, such as output labels or hints useful for evaluating system correctness, and a performance metric $\mu:\mathcal{A}\times\mathcal{M}\to\mathbb{R}$, the goal of compound AI system optimization is defined by solving:
\begin{equation}
\max_{\Phi} \frac{1}{N}\sum_{i=1}^N \mu\bigl(\Phi(q_i), m_i\bigr).    
\end{equation}

%% file: text/ai_system.tex
\section{Compound AI System Optimization}\label{sec:ai_system}
With background established, we now turn to the core of our survey. Sec.~\ref{sec:four_dim} first present four principled dimensions underlying the optimization of compound AI systems, i.e., \emph{Structural Flexibility}, \emph{Learning Signals}, \emph{Component Options}, and \emph{System Representations}. Sec.~\ref{sec:review} then review existing methods in the proposed 2×2 taxonomy.

\input{text/s3_1_four_dimension}

\subsection{Representative Methods}\label{sec:review}
Equipped with the discussed four principled dimensions, here we review existing methods along two major axes: \emph{Structural Flexibility} (Fixed vs. Flexible Structure) and \emph{Learning Signals} (NL Feedback vs. Numerical Signals)\footnote{For a few methods that use both types of learning signals, we still categorize them into a single category. See Appendix Sec.~\ref{sec:learning_details} for more details.}. All considered papers are listed in Table~\ref{tab:main}. 
\input{assets/main_table}

\input{text/s3_2_methods_review/NL_fixed}

\input{text/s3_2_methods_review/Numerical_fixed}

\input{text/s3_2_methods_review/NL_flexible}

\input{text/s3_2_methods_review/Numerical_flexible}

%% file: text/s3_1_four_dimension.tex

\begin{figure*}[t]
\centering
\includegraphics[width=0.90\textwidth, height=0.55\columnwidth]{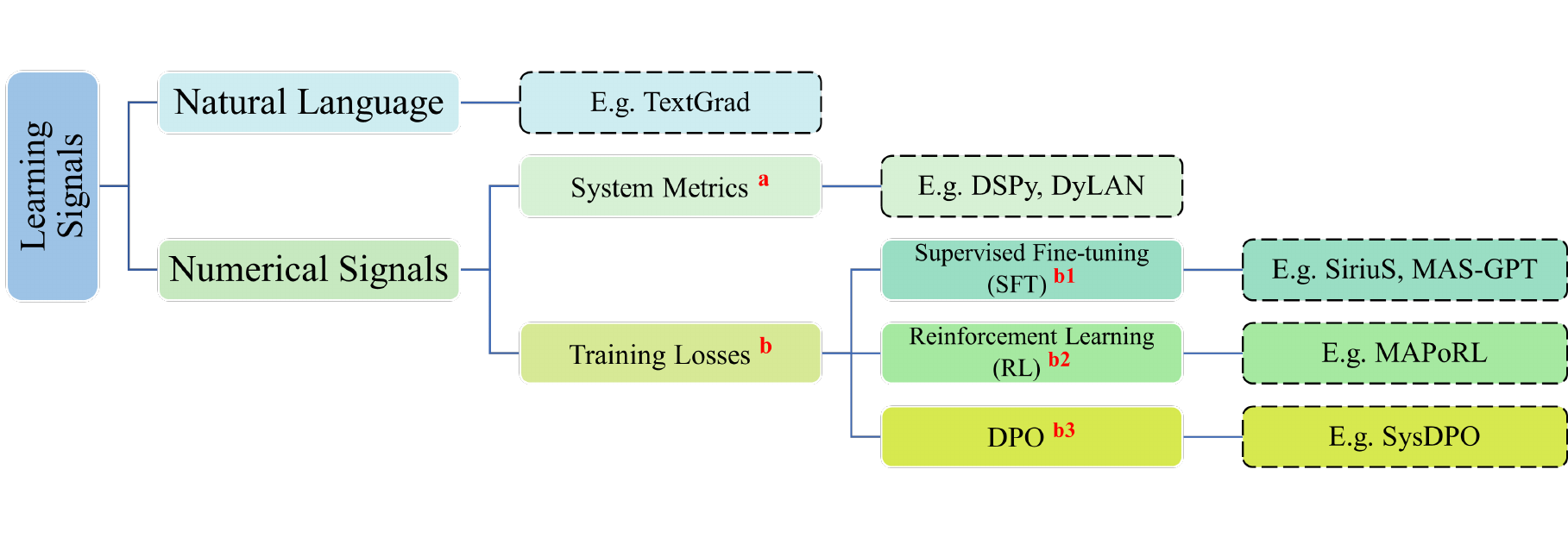}
\vspace{-2mm}
  \caption{\textbf{Learning Signals} are classified into two categories, with Numerical Signals further divided by their utilization schemes: (a) one class of methods devises rule-based algorithms that directly learn from raw system performance metrics, and (b) another class transforms system evaluation results into formalized training objectives. These objectives are further split as (b1) supervised fine-tuning (SFT) losses, (b2) reinforcement learning (RL) reward functions, and (b3) direct preference optimization (DPO)~\cite{DPO} losses.}
  \label{fig:learning_signals}
\end{figure*}

\subsection{Four Principled Dimensions}\label{sec:four_dim}
\paragraph{Structural Flexibility} 
As described in Sec.~\ref{sec:formal_def}, the pipeline of $\Phi$ can be modeled as a computation graph $G=(V,E)$. The concept of \emph{Structural Flexibility} thus refers to the degree to which an optimization method can modify this graph. One class of methods, termed \emph{Fixed Structure}, assumes a predefined topology $(V,E)$ and focuses exclusively on optimizing the node parameters $\{\Theta_i\}$ (e.g., LLM prompts). In contrast, more recent methods acknowledge the importance of identifying optimal system topologies~\cite{zhuge2024_gptswarm, hu2024_ADAS} and propose to jointly optimize both the node parameters and the graph structure itself, such as edge connections $E$, node counts $|V|$, and even the types of operations in $\mathcal{F}$. These methods, termed \emph{Flexible Structure}, broaden the search space of $\Phi$, enabling more effective orchestration of the system toward targeted tasks.

\paragraph{Learning Signals}
Regardless of structural flexibility, effective optimization of $\Phi$ requires~\emph{Learning Signals} that steer system updates toward improved task performance (Fig.~\ref{fig:sys_demo}). These signals naturally originate from system performance metric $\mu\bigr(\Phi(q_i), m_i\bigr)$, and can be conveyed in two distinct forms. The first type is natural language feedback (denoted as \emph{NL Feedback}), where signals are generated by an auxiliary LLM that is separate from those used within the system itself \cite{Mert2024_textgrad, cheng2024_trace, hu2024_ADAS}. This approach leverages the reasoning capabilities of LLMs to provide textual guidance for system optimization. The second type, termed~\emph{Numerical Signals}, encompasses methods in which learning signals are represented numerically. We further categorize numerical signals into four different forms, as illustrated in Fig.~\ref{fig:learning_signals}.

\paragraph{Component Options}
This dimension distinguishes optimization methods by their considered types of operations in $\mathcal{F}$. Though most $\Phi$ center around LLMs, many frameworks integrate additional components to enrich domain knowledge or perform specialized tasks. These include RAG modules~\cite{yin2025_llm_autodiff}, code interpreters (CI)~\cite{hu2024_ADAS, wang2025_scoreflow}, or Tools calling such as web search~\cite{zhuge2024_gptswarm}.
In multi-modal contexts, additional components such as image generation models (IGM) are employed~\cite{wang2025_systemDPO}. Note that some works assume an unrestricted components pool~\cite{Mert2024_textgrad} or do not explicitly specify their supported component options~\cite{hu2024_ADAS, wang2025_scoreflow}. In those cases, we infer their component options by examining the components used in their experimental setups or by inspecting their system representations.\footnote{Methods that use Python code as \emph{System Representations} inherently support code interpreters (CI).}

\paragraph{System Representations}
Different representations are used to characterize $\Phi$’s operational topology across methods. The most common choice is to model $\Phi$ as a graph. A directed acyclic graph (DAG) ensures that each node is invoked at most once per forward pass~\cite{khattab2023_dspy, zhuge2024_gptswarm}, while a cyclic graph supports schemes such as multi-turn debates~\cite{subramaniam2025multiagentFT, park2025_maporl} or multi-hop
RAG~\cite{yin2025_llm_autodiff}. Although~\citet{DyLAN} employs feed-forward networks as system representations, we also label them as graphs, given their equivalence. Acknowledging the potential limitations to optimize system topologies in graph space~\cite{hu2024_ADAS}, another line of research represents $\Phi$'s workflow as natural language program~\cite{li2024autoflow} or Python code~\cite{hu2024_ADAS, zhang2024_aflow, ye2025_masgpttrainingllmsbuild}, which supports conditional logic and loops without any acyclicity constraints.

%% file: assets/main_table.tex
\begin{table*}[ht]
    \centering
    \label{tab:optimized_order}
    \small
    \begin{tabular}{llllll}
    \toprule
    \multirow{2}{*}{\bf Methods} & \bf Structural & \bf Learning  & \bf Component & \bf System \\ 
    & \bf Flexibility & \bf Signals & \bf Options & \bf Representations\\
    \midrule
    \textbf{TextGrad}~\shortcite{Mert2024_textgrad}            & Fixed    & NL Feedback       & LLM         & Graph{\color{ForestGreen}$\mathlarger{\dagger}$}     &              \\ 
    \textbf{Trace}~\shortcite{cheng2024_trace}              & Fixed    & NL Feedback       & LLM         & Graph{\color{ForestGreen}$\mathlarger{\dagger}$}   \\ 
    \textbf{AIME}~\shortcite{patel2024_aime}                & Fixed    & NL Feedback       & LLM         & Graph{\color{ForestGreen}$\mathlarger{\dagger}$}    \\ 
    \textbf{Revolve}~\shortcite{zhang2024_revolve}            & Fixed    & NL Feedback       & LLM         & Graph{\color{ForestGreen}$\mathlarger{\dagger}$} \\ 
    \textbf{GASO}~\shortcite{wang2024correctly}   & Fixed    & NL Feedback       & LLM         & Graph{\color{ForestGreen}$\mathlarger{\dagger}$}  \\ 
    \textbf{LLM-AutoDiff}~\shortcite{yin2025_llm_autodiff} & Fixed    & NL Feedback       & LLM; RAG    & Graph    \\ 
    \midrule
    \textbf{DSPy}~\shortcite{khattab2023_dspy}                & Fixed    & Numerical Signals$^{\color{red}{\bm{a}}}$ & LLM & Graph{\color{ForestGreen}$\mathlarger{\dagger}$}  \\ 
    \textbf{MIPRO}~\shortcite{opsahl2024optimizing} & Fixed    & Numerical Signals$^{\color{red}{\bm{a}}}$ & LLM; RAG  & Graph \\ 
    \textbf{BetterTogether}{\color{blue}$*$}~\shortcite{soylu2024fine} & Fixed    & Numerical Signals$^{\color{red}{\bm{a, b1}}}$& LLM; RAG  & Graph{\color{ForestGreen}$\mathlarger{\dagger}$}   \\ 
    \textbf{SiriuS}{\color{blue}$*$}~\shortcite{zhao2025_sirius}              & Fixed    & Numerical Signals$^{\color{red}{\bm{b_{1}}}}$ & LLM  & Graph\\ 
    \textbf{MAPoRL}{\color{blue}$*$}~\shortcite{park2025_maporl}              & Fixed    & Numerical Signals$^{\color{red}{\bm{b_{2}}}}$ & LLM  & Graph\\ 
    \textbf{SysDPO}{\color{blue}$*$}~\shortcite{wang2025_systemDPO}    & Fixed    & Numerical Signals$^{\color{red}{\bm{b_{3}}}}$& LLM; IGM & Graph{\color{ForestGreen}$\mathlarger{\dagger}$}  \\ 
    \textbf{Multiagent Finetuning}{\color{blue}$*$}~\shortcite{subramaniam2025multiagentFT} & Fixed   & Numerical Signals$^{\color{red}{\bm{b_{1}}}}$& LLM  & Graph \\ 
    \midrule
    \textbf{Agent Symbolic Learning}~\shortcite{zhou2024_symbolic}  & Flexible & NL Feedback  & LLM; RAG  & Graph{\color{ForestGreen}$\mathlarger{\dagger}$}  \\ 
    \textbf{ADAS}~\shortcite{hu2024_ADAS} & Flexible & NL Feedback       & LLM; CI     & Python Code   \\ 
    \textbf{AFlow}~\shortcite{zhang2024_aflow} & Flexible & NL Feedback       & LLM; CI     & Python Code  \\ 
    \textbf{MASS}~\shortcite{zhou2025_multi}& Flexible & NL Feedback       & LLM      & Graph   \\ 
    \textbf{DebFlow}~\shortcite{su2025debflowautomatingagentcreation}             & Flexible & NL Feedback       & LLM     & Graph \\ 
    \midrule
    \textbf{DyLAN}~\shortcite{DyLAN} & Flexible & Numerical Signals$^{\color{red}{\bm{a}}}$ & LLM     & Graph  \\ 
    \textbf{GPTSwarm}~\shortcite{zhuge2024_gptswarm}  & Flexible & Numerical Signals$^{\color{red}{\bm{b_{2}}}}$& LLM; Tools  & Graph{\color{ForestGreen}$\mathlarger{\dagger}$}  \\ 
    \textbf{AutoFlow}{\color{blue}$*$}~\shortcite{li2024autoflow}            & Flexible & Numerical Signals$^{\color{red}{\bm{b_{2}}}}$& LLM         & NL Programs \\ 
    \textbf{MaAS}~\shortcite{zhang2025multi}            & Flexible & Numerical Signals$^{\color{red}{\bm{b_{2}}}}$& LLM; Tools; CI     &  Graph{\color{ForestGreen}$\mathlarger{\dagger}$} \\ 
    \textbf{ScoreFlow}{\color{blue}$*$}~\shortcite{wang2025_scoreflow}            & Flexible & Numerical Signals$^{\color{red}{\bm{b_{3}}}}$& LLM; CI     & Python Code \\ 
    \textbf{MAS-GPT}{\color{blue}$*$}~\shortcite{ye2025_masgpttrainingllmsbuild}         & Flexible & Numerical Signals$^{\color{red}{\bm{b_{1}}}}$ & LLM; CI     & Python Code  \\ 
    \textbf{W4S}{\color{blue}$*$}~\shortcite{nie2025weak} & Flexible & Numerical Signals$^{\color{red}{\bm{b_{2}}}}$& LLM; CI     & Python Code  \\ 
    \textbf{FlowReasoner}{\color{blue}$*$}~\shortcite{gao2025flowreasoner} & Flexible & Numerical Signals$^{\color{red}{\bm{b_{1}},\bm{b_{2}}}}$& LLM; CI     & Python Code  \\ 
    \bottomrule
    \end{tabular}
    \caption{\textbf{Compound AI System Optimization methods}, sorted by their first publication date on arXiv. All methods and their properties along the four principled dimensions are listed. Superscripts $^{\color{red}{\bm{a}}}$, $^{\color{red}{\bm{b_{1}}}}$, $^{\color{red}{\bm{b_{2}}}}$, and $^{\color{red}{\bm{b_{3}}}}$ denote the type of numerical signal each method employs (Fig.~\ref{fig:learning_signals}). An asterisk ({\color{blue}$*$}) indicates methods that require model fine-tuning. For graph-based system representations, a dagger ({\color{ForestGreen}${\mathlarger{\dagger}}$}) marks methods restricted to acyclic structures (i.e., DAGs).}
    \label{tab:main}
    \vspace{-3mm}
\end{table*}

%% file: text/s3_2_methods_review/NL_fixed.tex
\paragraph{Fixed Structure, NL Feedback}
Most existing methods leverage LLMs' ability to generate rich, general natural language suggestions for prompt optimization in standalone LLMs~\cite{APE, APO, OPRO}. TextGrad~\cite{Mert2024_textgrad} is among the first to extend this idea to compound AI systems by updating their node parameters, e.g., LLM prompts ($\{\theta_{i,T}\}$). Drawing inspiration from numerical gradient descent in PyTorch~\cite{paszke2019pytorch}, TextGrad defines each node in the system graph as an independent computational unit, where optimization unfolds in three stages: (1) an \emph{evaluator} LLM assesses the system’s output $\Phi(q_i)$ against an expected reference $m_i$ and generates textual loss signals; (2) for each participating node, a \emph{gradient estimator} LLM generates node-specific textual suggestions conditioned on system dialogue and backpropagated loss; and (3) an \emph{optimizer} LLM at each node refines node parameters using these suggestions. This process mirrors backpropagation via natural language, simulating differentiability across discrete modules.

Several works have since built upon TextGrad’s framework. AIME~\cite{patel2024_aime} shows that for complex code generation tasks, using a single \emph{evaluator} LLM often allows errors in the generated code to go undetected, whereas concatenating outputs from multiple~\emph{evaluator} LLMs can mitigate this issue. REVOLVE~\cite{zhang2024_revolve} observes that \emph{NL Feedback} is often applied in a first-order manner, causing stagnation and oscillations during system optimization. It therefore enriches the \emph{gradient estimator} LLMs’ input with a concise execution history of past prompts and responses, enabling the generation of curvature-aware feedback, similar to the Hessian in numerical optimization. GASO~\cite{wang2024semantic_backpropagation} identifies the negligence of sibling-input interactions in TextGrad’s backpropagation scheme, thus proposes semantic gradient descent to compute context-aware gradients and aggregate them for credit assignment. LLM-AutoDiff~\cite{yin2025_llm_autodiff} addresses the intricacy of multi-component (e.g., large $|V|$, diverse $\mathcal{F}$) and cyclic system structures that prior work has not fully explored. In particular, it introduces \emph{time-sequential gradients} to accumulate multiple textual gradients for nodes invoked repeatedly during a forward pass, and proposes an optional \emph{skip-connections} mechanism, serving as powerful building block for optimizing large-scale systems via natural language. 

Trace~\cite{cheng2024_trace}, developed concurrently with TextGrad, introduces a joint optimization that updates all LLM prompts at once. It first obtains global \emph{NL Feedback} from an~\emph{evaluator} LLM and then, in a single LLM invocation, updates every involved node by presenting the model with an minimal subgraph (analogous to execution trace in Python). This process address two caveats~\cite{lin2024_llm_opt} in the backpropagation scheme of TextGrad: (i) error accumulation due to imprecise \emph{NL Feedback}, and (ii) the linear growth in LLM calls with the number of nodes. 

Serving as pioneers in the field, these methods demonstrate that learning from text is possible not only for single AI models but also for systems composed of discrete modules. Nevertheless, their successes are supported mainly by empirical results without theoretical grounding. In addition, the common reliance on proprietary LLMs to generate NL feedback incurs high API costs. 

%% file: text/s3_2_methods_review/Numerical_fixed.tex
\paragraph{Fixed Structure, Numerical Signals}
Rather than relying on textual signals, methods in this category use numerical signals to update node parameters. DSPy~\cite{khattab2023_dspy} provides a Python library\footnote{\url{http://dspy.ai}} featuring declarative programming modules for designing and optimizing $\Phi$. As a method that learns directly from raw system metrics, it introduces a suite of rejection‐sampling–based routines (\texttt{Bootstrap-*}) for generating high‐quality in‐context demonstrations informed by corresponding system performance. Users may optionally fine‐tune LLM weights ($\theta_{i,N}$) on the collected demonstrations via \texttt{BootstrapFinetune} procedure. MIPRO~\cite{opsahl2024optimizing} advances by jointly optimizing demonstrations and instructions. Specifically, it employs a Bayesian surrogate model to maintain and update posterior distributions over instruction–demonstration configurations, favoring those that yield high performance. BetterTogether~\cite{soylu2024fine} extends by alternating between LLM prompt and weight optimization, enabling LLMs to iteratively teach themselves and outperform single-strategy baselines.

Remaining methods in this category involve model fine-tuning, where numerical signals from system evaluation are instantiated as different types of training losses, as discussed in Fig.~\ref{fig:learning_signals}. In the SFT category, SiriuS~\cite{zhao2025_sirius} constructs several system schemes by assigning predefined roles to its LLM nodes (e.g., “Physicist” and “Mathematician”). It then gathers reasoning trajectories, i.e., intermediate dialogue outputs, for queries \(q_i\) with high \(\mu\bigl(\Phi(q_i), m_i\bigr)\) and independently supervised fine-tunes each LLM node using its corresponding input–output pairs. For \(q_i\) that results in failed attempts, SiriuS performs~\emph{trajectory augmentation} by resampling original attempts
with feedback from an additional agent. Concurrent with SiriuS but building on the multiagent debate~\cite{du2023improving} scheme, \emph{multiagent finetuning}~\cite{subramaniam2025multiagentFT} introduces novel rules for collecting training data for each \emph{generation} and \emph{critic} model for subsequent SFT. 

Belonging to the RL category, MAPoRL~\cite{park2025_maporl} targets the multiagent debate~\cite{du2023improving} scenario as well. It differs from ~\citet{subramaniam2025multiagentFT} by training a~\emph{verifier} LLM to assign immediate correctness rewards to each LLM node, and introduces influence-aware reward shaping to incentivize collaboration. Finally, in the DPO category, SysDPO~\cite{wang2025_systemDPO} features a system characterized by an LLM and a diffusion model~\cite{rombach2022high}. Aiming at image generation tasks, SysDPO curates a preference dataset by computing preference scores based on images’ \emph{order consistency} and \emph{distribution evenness}. Unlike original DPO loss, SysDPO incorporates probability decomposition to enable fine-tuning multiple components in $\Phi$.

These methods introduce novel ways to leverage system performance signals, effectively mitigating the imprecision of natural language. However, excessive human‐designed rules may limit generalizability. Additionally, fine‐tuning each LLM in \(\Phi\) incurs substantial GPU resource demands.

%% file: text/s3_2_methods_review/NL_flexible.tex
\paragraph{Flexible Structure, NL Feedback}
Methods discussed previously tune only node parameters within a predefined system topology. To overcome this constraints, methods in this category leverage NL Feedback to jointly optimize both node parameters and overall system structure. Concurrent with TextGrad~\cite{Mert2024_textgrad}, Agent Symbolic Learning~\cite{zhou2024_symbolic} designs optimizers with three components:~\emph{PromptOptimizer},~\emph{ToolOptimizer}, and~\emph{PipelineOptimizer}, making it goes beyond node tuning because the latter two components support tool creation as well as node addition, deletion, and movement.
Through a series of analytical experiments, MASS~\cite{zhou2025_multi} demonstrates that optimizing LLM prompts $\{\theta_{i,T}\}$ can deliver performance gains in compound AI systems more easily than exploring the topology $(V, E)$. Building on this insight, MASS devises a three‐stage framework in which prompt optimization precedes topology search.

Recognizing that previous approaches using graphs as system representations may not fully cover the space of possible system designs and that this space is inherently difficult to search, ADAS~\cite{hu2024_ADAS} is the first to adopt Python code as system representation. 
Specifically, conditioned on an archive maintaining prior code and its corresponding performance metric, a meta LLM is asked to iteratively design novel workflows. Owing to the vast search space of code representations, ADAS faces challenges in its linear heuristic search processes and coarse workflow storage, which lead to the accumulation of irrelevant information for the meta LLM~\cite{zhang2024_aflow}. Identifying this, AFlow~\cite{zhang2024_aflow} leverages the tree structure of Monte Carlo Tree Search (MCTS) to preserve past experience and employs a set of predefined operators to efficiently identify optimal system designs. DebFlow~\cite{su2025debflowautomatingagentcreation} argues the limitation posed by using a single meta LLM in this scheme, and introduces an innovative debate framework where multiple~\emph{debaters} propose their opinions on system designs, with a~\emph{judge} helping to conclude the refined workflow.

Methods in this category demonstrate how novel algorithms can leverage the power of proprietary LLMs for direct workflow design. Nonetheless, effective engagement with these meta LLMs demands high token consumption~\cite{hu2024_ADAS} or multiple LLM inferences~\cite{su2025debflowautomatingagentcreation}. Also, evaluations on less powerful open-source models remain scarce.

%% file: text/s3_2_methods_review/Numerical_flexible.tex
\paragraph{Flexible Structure, Numerical Signals}
Completing our review, this passage describes methods that employ \emph{Numerical Signals} to refine $\Phi$ without posing fixed topology constraints. DyLAN~\cite{DyLAN} and GPTSwarm~\cite{zhuge2024_gptswarm} are methods in this category that do not involve model fine-tuning. DyLAN models multiturn debate as a temporal feed-forward network in which agents prompted with distinct roles are unrolled across layers. The system is optimized through pruning unhelpful agents and adaptively connect the surviving agents between consecutive layers. Rule-based algorithms are employed to score each agent’s importance, positioning DyLAN within the category that leverages raw system metrics as learning signals. GPTSwarm models $\Phi$ as a hierarchical framework composed of \emph{nodes} ($v_i \in V$), \emph{agents} (graphs that link nodes), and \emph{swarms} (composite graphs that interconnect multiple agents), then employs an optimization procedure to refine the edge connections among \emph{agents}. A parameterized probabilistic distribution $D_{\theta}$ is introduced to govern the connectivity within the swarm graph, which is then optimized using a gradient‐ascent variant of the REINFORCE~\cite{williams1992simple} algorithm, placing GPTSwarm within the class of RL methods. 

Before discussing the remaining methods, we note that they operate in a query‐adaptive manner, i.e., instantiating a new \(\Phi\) for each \(q_i\). This operating scheme contrasts with methods discussed so far in which a single task‐specific \(\Phi\), once optimized, is used for all \(q_i\). Similar to ADAS~\cite{hu2024_ADAS} and AFlow~\cite{zhang2024_aflow}, these methods leverage a meta LLM to generate optimal system designs. However, rather than relying solely on inference, they fine-tune the meta LLM using constructed training datasets or reward functions to enable effective pipeline generation, given the limited prior knowledge about compound AI system design within LLMs

Starting from the SFT category, MAS-GPT~\cite{ye2025_masgpttrainingllmsbuild} first constructs a \emph{query pool} from open‐source queries across various domains, and a \emph{MAS pool} by manually implementing over 40 common system designs
in Python code. Through \emph{evaluation}, \emph{selection}, and \emph{refinement} of query–MAS pairs, MAS-GPT curates a training dataset that ensures consistency, minimizing ambiguity when fine‐tuning the meta LLM.

Next, within the RL category, AutoFlow~\cite{li2024autoflow} prompts the meta LLM to generate $\Phi$ using CoRE~\cite{xu2024core} syntax, then iteratively fine-tunes it
with the average score on task data $\mathcal{D}$ as the reward. It also offers a workaround for closed-source meta LLMs by in-context learning. MaAS~\cite{zhang2025multi} constructs and optimizes an \emph{agentic supernet}, a probabilistic
distribution over agentic architectures. The token cost incurred during system execution is also considered as a loss term to achieve a trade-off between system performance and complexity. W4S~\cite{nie2025weak} maximizes the meta LLM’s flexibility by constraining only the workflow interfaces, without predefining any system modules. By using a small (weak) model to reduce training cost, W4S casts the problem as a multi-turn Markov Decision Process (MDP), in which the meta LLM progressively learns to design and refine \(\Phi\) based on environmental feedback. FlowReasoner~\cite{gao2025flowreasoner} combines SFT and RL by first fine-tuning the meta LLM for basic reasoning regarding system generation, and then applying RL with a multi-purpose reward to further optimize the model. 

Finally, ScoreFlow~\cite{wang2025_scoreflow} introduces Score-DPO, an extension of the original DPO.
During each iteration, for every query $q_i$, multiple candidate $\Phi$ are sampled from the meta LLM and evaluated by an LLM executor; preference data are then collected based on the observed differences in system performance.

Recent trends in this category  frame compound system optimization as a meta LLM fine‐tuning problem, thereby sidestepping the non‐differentiability of modules within \(\Phi\). However, substantial effort is still required to curate high‐quality training data, and the lack of comprehensive evaluation across different model families limits the practical applicability of these methods. 

%% file: text/discussions.tex
\section{Challenges and Future Directions}\label{sec:discussions}

After reviewing representative methods, this section presents several key challenges and their future directions. Due to space constraints, additional discussion is deferred to Appendix~Sec.~\ref{sec:advanced}.

\paragraph{Manual Hyperparameter Configuration} 
Despite aiming to automate the optimization process, we identify residual human interventions in existing methods, particularly in the configuration of algorithm-related hyperparameters, that challenge automation claims and limit practical value.

In particular, methods in the Fixed Structure category require users to configure system topologies based on domain expertise. Although some studies evaluate their algorithms across multiple system designs~\cite{yin2025_llm_autodiff,zhao2025_sirius}, there is no guarantee that these configurations will meet the requirements of all target applications. Textual hyperparameters also frequently appear in various methods. For example, the prompt templates used for the \emph{evaluator}, \emph{gradient estimator}, and \emph{optimizer} in TextGrad~\cite{Mert2024_textgrad}, as well as those for the meta LLM in ADAS~\cite{hu2024_ADAS}, are handcrafted by the original authors and often lack a clear design rationale or sensitivity analysis of their wording.

Numerical hyperparameter decisions persist as well; for instance, the number of bootstrap samples in DSPy~\cite{khattab2023_dspy} remains user-tunable and cannot be automated. Even seemingly automated numerical flexible-structure methods, such as MAS-GPT~\cite{ye2025_masgpttrainingllmsbuild}, require manual configuration, as evidenced by the prompt templates for \emph{pair refinement}.

Despite the efforts of metaTextGrad~\cite{xu2024metatextgrad}, which applies meta‐learning to automatically optimize templates for \emph{evaluator}, human intervention remains to craft the meta‐learner’s prompts. To move toward truly automated system optimization, akin to near hyperparameter‐free neural network training, we urge future research to reduce reliance on both textual and numerical hyperparameters. For any hyperparameters that remain, we advocate thorough sensitivity analyses to help users understand each method’s behavior and robustness.


\begin{figure}
    \centerline{\includegraphics[width=1.0\columnwidth]{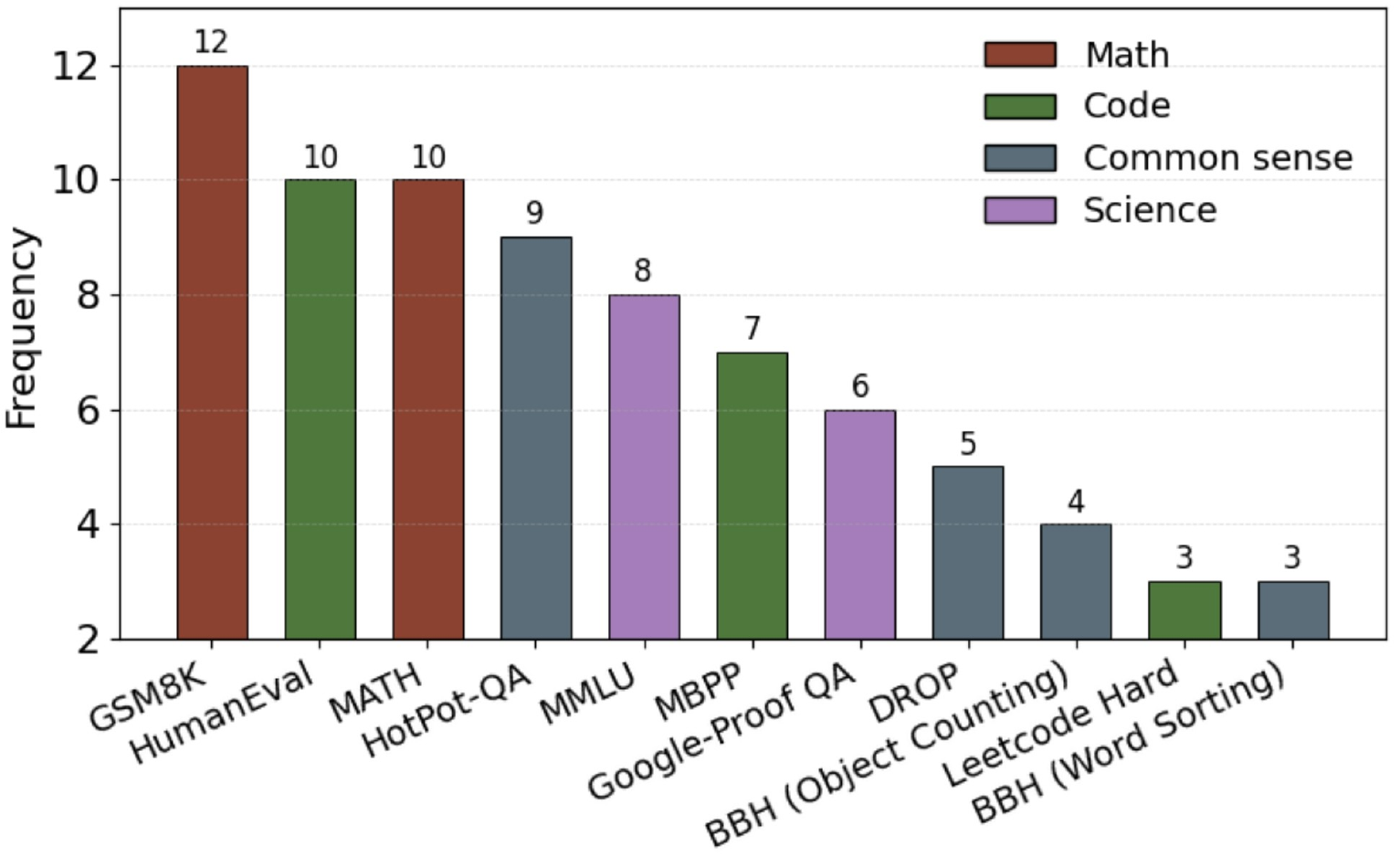}}
    \caption{The frequency statistics of benchmarks tested in the surveyed 26 papers.}
    \label{fig: example}

\end{figure}

\paragraph{Excessive Computation Burden}
Optimization of compound AI systems
is inherently more challenging than tuning individual models. Existing approaches thus resort to various workarounds, leading to substantially higher computational overhead.

In NL feedback learning, methods such as TextGrad require multiple LLM calls to approximate a single gradient step. 
Although methods like Trace~\cite{cheng2024_trace} and ADAS~\cite{hu2024_ADAS} use global LLM call(s) per optimization step, they must embed extensive context in their prompts (e.g.,~\emph{minimal subgraphs} or \emph{agent archives}), which increases token throughput. Since these methods typically rely on proprietary models~\cite{achiam2023gpt}, they incur substantial API costs. 
Conversely, numerical signal-based methods often leverage open-source LLMs to avoid API expenses. These models typically require fine-tuning to perform well, thereby shifting the burden to GPU resources. Developers are thus faced with a trade-off between API costs and GPU consumption.

Moreover, computational burden also arises during inference. By focusing primarily on system performance, current Flexible Structure methods often overlook the need to regularize system complexity (e.g., multi-round loops or lengthy executions), resulting in unbounded resource consumption at run time. Although a few methods~\cite{zhang2025multi, gao2025flowreasoner} have begun to encourage less complex system designs (e.g., lower token consumption), their applicability and scalability in large-scale deployments remain to be tested~\cite{liu2025advancesagents}. We therefore suggest that future research develop resource-efficient optimization algorithms and at the same time devise principled ways to constrain system complexity without compromising performance.
 
\paragraph{Limited Experimental Scope}
Since compound AI systems are expected to address complex problems, it is important to investigate their effectiveness on more challenging tasks. Yet, papers in the field primarily evaluate their proposed methods on datasets widely used for single LLMs (Fig.~\ref{fig: example}), such as those for math reasoning (e.g., GSM8K~\cite{cobbe2021trainingverifierssolvemath}), commonsense reasoning (e.g., MMLU~\cite{hendrycks2021measuringmassivemultitasklanguage}), and code generation (e.g., HumanEval~\cite{chen2021evaluating} and MBPP~\cite{austin2021program}). While these evaluations reflect general effectiveness, we argue that it is also important to include benchmarks involving more complex tasks. For instance, AgentBench~\cite{liu2023agentbench} and AgentGym~\cite{xi2024agentgym} comprise multiple tasks requiring different LLMs within the system to cooperate and discuss, and GAIA~\cite{mialon2023gaia} examines system effectiveness in real-world scenarios requiring various tool usage. Furthermore, given the broad utility of compound AI systems (e.g., healthcare systems with doctors embedded as nodes in $\Phi$~\cite{chen2024towards}), it is necessary to evaluate algorithmic performance when humans function as nodes within the system and to devise principled methods for modeling their behavior.

\paragraph{Empirical NL Feedback} 
While NL Feedback methods have shown promising empirical results, they lack theoretical guarantees. For example, the convergence of textual gradient descent remains unproven, whereas classical gradient descent are supported by formal convergence proofs~\cite{cheridito2022proof,hutzenthaler2021convergence}.
Such proofs provide a solid foundation for the continuous advancement of single‐model optimization. We therefore advocate future work to deliver rigorous convergence and optimality analyses for learning via NL Feedback, which will offer deeper insights and strengthen the field’s theoretical underpinnings. 

\paragraph{Potential Safety Risks} 
While safety issues and corresponding defenses for single models have been extensively studied, such as jailbreak attacks on LLMs and their mitigation~\cite{yi2024jailbreak} and human-engineered AI pipelines to reduce harmful prompts and outputs~\cite{han2024torchopera}, the attack surface expands considerably in compound AI systems~\cite{banerjee2024sok}. For instance, \citet{debenedetti2024privacy} demonstrate that privacy-preserving models can still leak sensitive data when integrated as a component of a larger system. Furthermore, because compound AI systems are often embodied and executed as code within enterprise environments, latent failure modes may remain undetected and undermine system reliability even in the absence of explicit adversarial attacks. Despite these risks, research on compound AI system optimization that extends beyond downstream performance has so far addressed mainly execution efficiency~\cite{zhang2025multi}, with little attention to system-level alignment or safety~\cite{zheng2025mermaidflow}. Given the mature alignment and safeguarding optimization techniques available for single models~\cite{achiam2017constrained, dai2023safe}, we urge future work to adapt and extend these strategies to compound systems in order to balance capability enhancements with safety guarantees~\cite{yang2024ai45circlawroadmaptrustworthy}.

\paragraph{Inconsistent Library Support}
During our survey, we observed a lack of a standardized and widely adopted library in the field. Although some well-maintained libraries such as TextGrad~\cite{Mert2024_textgrad} and DSPy~\cite{khattab2023_dspy} have gained popularity among practitioners, still a great portion of existing works implement compound AI systems optimization using custom, self-crafted codebases.
While frameworks such as TensorFlow~\cite{Abadi_TensorFlow_Large-scale_machine_2015} and PyTorch~\cite{paszke2019pytorch} dominate single-model training, best practices for implementing and optimizing compound AI systems are still under development.
We suggest that future efforts focus on systematically benchmarking and comparing existing libraries for compound AI system optimization. Such efforts could support their improvement and help establish clearer guidelines for developers and researchers—for example, regarding when to use which library—as is done in the context of single-model training~\cite{novac2022analysis}.

%% file: text/conclusion.tex
\section{Conclusions} \label{sec:conclusion}
We survey recent advances in optimizing compound AI systems composed of interacting components like agents and tools. 
To unify diverse research efforts, we propose a graph-based formalism with conditional edges, enabling structured analysis of system interactions. With this framework, we examine existing methods across four principled dimensions and organize them into a 2$\times$2 taxonomy. Our survey highlights key trends and trade-offs, including computational overhead, the NL interface, and challenges in scaling and generalization. We also identify open problems and outline future directions to guide continued progress in the field.




%% file: appendix/A1.tex
\section{More Details on Formal Definitions}\label{sec:more_math}
\paragraph{Special Nodes} In our formalism of compound AI systems, two special nodes play key roles: 

\noindent (1) the input node \(v_{\mathrm{in}}\in V\), which receives the user query \(q\in\mathcal{Q}\) as the system’s entry point; and 

\noindent (2) the output node \(v_{\mathrm{out}}\in V\), which terminates system execution. 
In our definition, the operation attached to \(v_{\mathrm{out}}\) is the identity function \(I\), so that it simply parses its input as the final system answer:
\begin{gather*}
  Y_{\mathrm{out}} = I\bigl(X_{\mathrm{out}}\bigr)\,,\\
  a = Y_{\mathrm{out}}\,.
\end{gather*}
Additionally, all outgoing conditional edges from $v_{\text{out}}$ are set to zero ($c_{\text{out},j} = 0, \forall j$), ensuring that the system does not route its output back to other nodes. This terminal structure effectively signals computation completion, similar to how end-of-sequence tokens function in language models.

\paragraph{Support for Cyclic Structures}  
Our formulation naturally accommodates both DAGs and cyclic topologies via conditional edges. Formally, a cycle at node \(v_i\) exists if and only if there is a path of length \(L\ge1\)  
\[
\bigl(v_{i_0},v_{i_1},\dots,v_{i_L}\bigr)\subseteq V
\]
such that  
\[
v_{i_0}=v_{i_L}=v_i,
\quad
v_{i_t}\neq v_{\text{out}}\ \ \forall\,0\le t<L,
\]
and
\[
c_{i_t,i_{t+1}}(q,\tau_t)=1
\quad
\forall\,0\le t< L.
\]
Here, \(\{v_{i_0},\dots,v_{i_L}\}\) denotes the sequence of nodes visited along the loop, and each \(c_{i_t,i_{t+1}}(q,\tau_t)=1\) indicates that the conditional edge from \(v_{i_t}\) to \(v_{i_{t+1}}\) is active under input \(q\) and state \(\tau_t\). In other words, \(v_i\) can route its output back to itself before reaching the output node \(v_{\text{out}}\), forming a cyclic loop.

If one would like to enforce acyclicity, it suffices to record the history of visited nodes in the state \(\tau\) and disallow any transition \(v_j\to v_k\) whenever \(v_k\) already appears in the history. This simple rule  easily prevents the re-entry required to form a cycle.

\begin{figure}[ht]
  \centering
  \begin{subfigure}[b]{0.48\columnwidth}
    \centering
    \includegraphics[width=\linewidth]{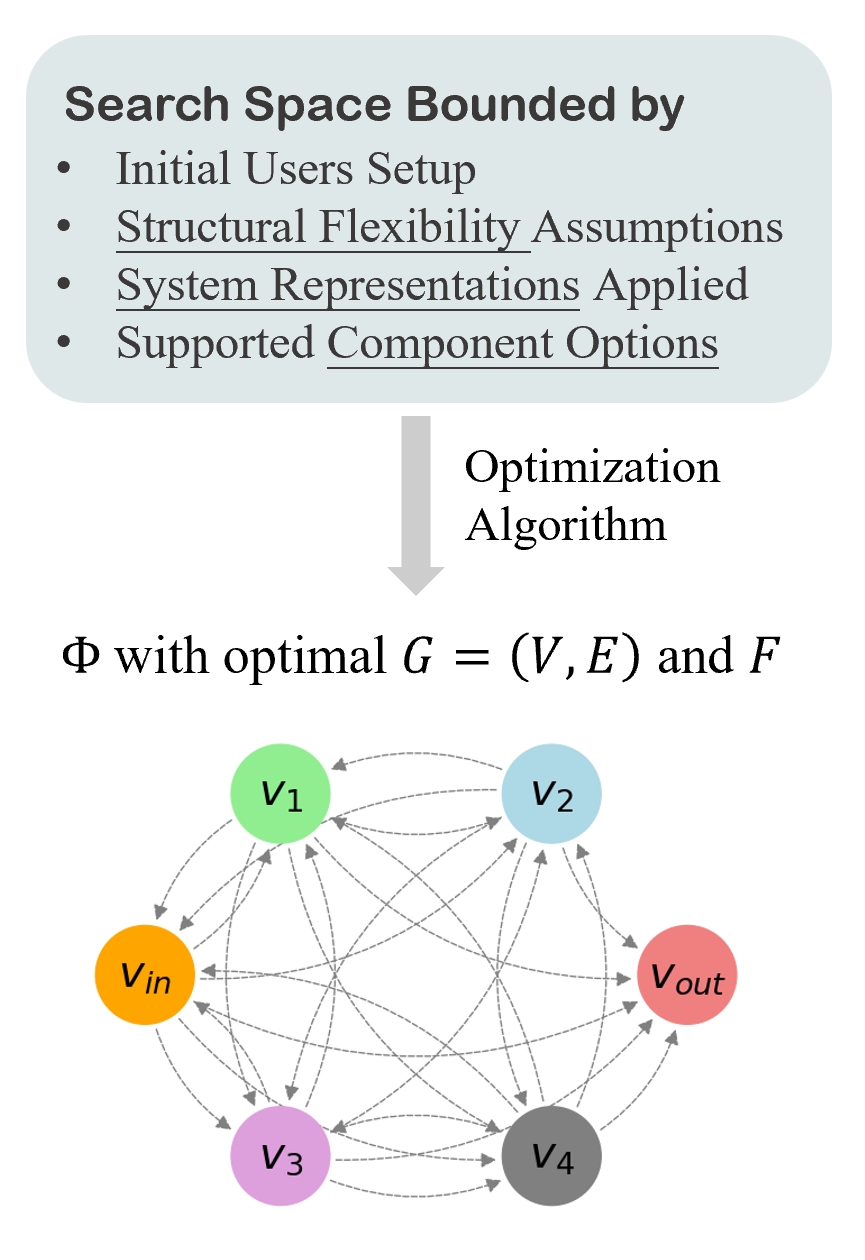}
    \caption{Optimization Stage}
    \label{fig:panel_a}
  \end{subfigure}
  \hfill
  \begin{subfigure}[b]{0.48\columnwidth}
    \centering
    \includegraphics[width=\linewidth]{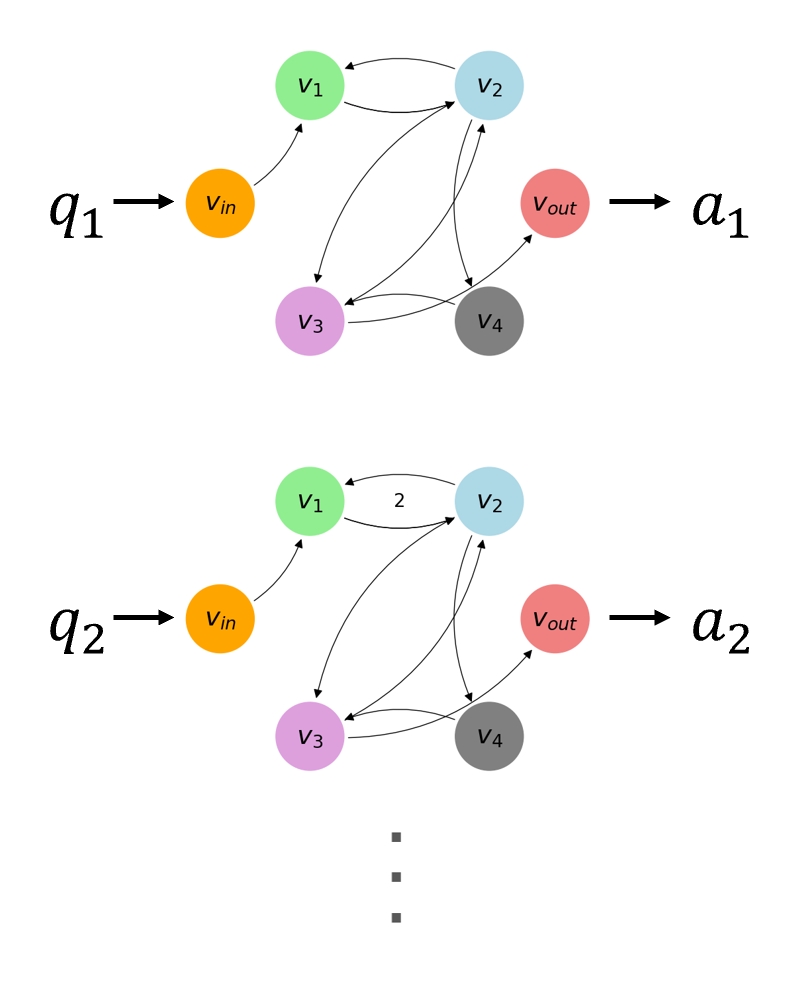}
    \caption{Execution Stage}
    \label{fig:panel_b}
  \end{subfigure}
  \caption{(a) During optimization, the algorithm explores the design space defined by user constraints and algorithmic parameters (Sec.~\ref{sec:four_dim}). Although the conditional arguments in the edge matrix $E = [c_{ij}]$ are fixed once optimization completes, the actual on/off status of each $c_{ij}$ remains undetermined. (b) At runtime, the optimized $\Phi$ instantiates different execution topologies based on $c_{ij}(\tau)$, reflecting dependence on the query input $q_i$ and the induced contextual state $\tau$.}
  \label{fig:ai_system}
\end{figure}

\paragraph{Visualization} Fig.~\ref{fig:ai_system} provides a visualization of system optimization under our formalism of graphs with conditional edges.

%% file: appendix/A2.tex
\section{Advanced Topics and Applications}\label{sec:advanced}
We briefly highlight several advanced topics and emerging applications in compound AI systems that may interest readers. Serving as advanced optimization methods, LLMSelector~\cite{chen2025optimizing} shows that selecting different LLMs at system nodes has a major effect on system performance, proposing an efficient algorithm for model selection in compound systems. Efforts have also focused on improving system execution efficiency, including network orchestration techniques~\cite{santhanam2024alto} and hardware-level optimizations~\cite{chaudhry2025towards}. These strategies are particularly well suited for refining complex system pipelines, such as question answering with RAG or data processing with tool calling, by accounting for the run-time efficiency.

The advances in compound AI systems have also led to novel applications: SciAgents~\cite{ghafarollahi2024sciagents} leverage ontological knowledge graphs and data retrieval tools to drive automatic materials discovery that surpasses traditional human-driven research methods; AutoML-Agent~\cite{trirat2024autoML} devises a system to handle end-to-end AutoML pipeline, accepting user task descriptions and optional constraints and handling everything from data retrieval to model deployment; and VFlow~\cite{wei2025vflow} targets hardware design tasks (i.e., Verilog code generation) by augmenting AFlow~\cite{zhang2024_aflow} with domain-specific components, thereby optimizing system performance to achieve a higher pass@1 rate than previous methods.

Furthermore, the effectiveness of natural language feedback demonstrated by TextGrad~\cite{Mert2024_textgrad} has spurred applications in diverse domains. For example, FedTextGrad~\cite{chen2025fedtextgrad} investigates the potential and challenges of integrating TextGrad into federated learning settings, where the server receives and aggregates locally optimized prompts from clients. Similarly, TPO~\cite{li2025TPO} proposes a method for aligning LLM preferences during inference by leveraging textual gradient signals. 

\section{More Details of Learning Signals}\label{sec:learning_details}
Most surveyed papers leverage a single type of learning signal (i.e., either NL feedback or numerical signals) to guide the optimization of $\Phi$. However, several works use both types of signals. For instance, GPTSwarm~\cite{zhuge2024_gptswarm} employs an RL loss to numerically learn optimal connections within the swarm and then updates LLM prompts using OPRO~\cite{OPRO}, an algorithm based on NL feedback. Additionally, MaAS~\cite{zhang2025multi} leverages a Bayesian Monte Carlo procedure to numerically update the probability distribution of the \emph{agentic supernet} and uses textual gradients to refine its \emph{operators}, including LLM prompts, temperature settings, and local node structures.

Despite the existence of these methods, we do not introduce a separate “hybrid” category for now. Instead, we classify them as either NL feedback or numerical signals based on two criteria: (1) the authors’ primary design novelty—for example, GPTSwarm is categorized as numerical since OPRO is a previously designed algorithm; and (2) the signal used to update the system topology—for example, MaAS is categorized as numerical since the topology is updated via numerical signals.

To accommodate future developments, we will dynamically adjust our criteria to keep the taxonomy clear and intuitive. Introducing a “hybrid” category also remains an open option as the field evolves. We encourage future efforts that leverage both types of learning signals to leverage our proposed taxonomy and position their methods at the intersection of NL feedback and numerical signals.

\section{Clarification of Methods Name}
Most surveyed papers introduce a single name for their method (i.e., main algorithm), while others employ multiple terms to refer to the target problem~\cite{wang2024correctly, hu2024_ADAS}, individual algorithms or components, or released libraries~\cite{khattab2023_dspy}. To improve readability, we assign these works a single, consistent name, chosen based on (1) its prevalence in subsequent literature and (2) its appearance in the original paper’s title.

%% file: main.bbl
\begin{thebibliography}{84}
\providecommand{\natexlab}[1]{#1}

\bibitem[{Abadi et~al.(2015)Abadi, Agarwal, Barham, Brevdo, Chen, Citro, Corrado, Davis, Dean, Devin, Ghemawat, Goodfellow, Harp, Irving, Isard, Jozefowicz, Jia, Kaiser, Kudlur, Levenberg, Mané, Schuster, Monga, Moore, Murray, Olah, Shlens, Steiner, Sutskever, Talwar, Tucker, Vanhoucke, Vasudevan, Viégas, Vinyals, Warden, Wattenberg, Wicke, Yu, and Zheng}]{Abadi_TensorFlow_Large-scale_machine_2015}
Martín Abadi, Ashish Agarwal, Paul Barham, Eugene Brevdo, Zhifeng Chen, Craig Citro, Greg~S. Corrado, Andy Davis, Jeffrey Dean, Matthieu Devin, Sanjay Ghemawat, Ian Goodfellow, Andrew Harp, Geoffrey Irving, Michael Isard, Rafal Jozefowicz, Yangqing Jia, Lukasz Kaiser, Manjunath Kudlur, Josh Levenberg, Dan Mané, Mike Schuster, Rajat Monga, Sherry Moore, Derek Murray, Chris Olah, Jonathon Shlens, Benoit Steiner, Ilya Sutskever, Kunal Talwar, Paul Tucker, Vincent Vanhoucke, Vijay Vasudevan, Fernanda Viégas, Oriol Vinyals, Pete Warden, Martin Wattenberg, Martin Wicke, Yuan Yu, and Xiaoqiang Zheng. 2015.
\newblock \href {https://doi.org/10.5281/zenodo.4724125} {{TensorFlow, Large-scale machine learning on heterogeneous systems}}.

\bibitem[{Achiam et~al.(2023)Achiam, Adler, Agarwal, Ahmad, Akkaya, Aleman, Almeida, Altenschmidt, Altman, Anadkat et~al.}]{achiam2023gpt}
Josh Achiam, Steven Adler, Sandhini Agarwal, Lama Ahmad, Ilge Akkaya, Florencia~Leoni Aleman, Diogo Almeida, Janko Altenschmidt, Sam Altman, Shyamal Anadkat, et~al. 2023.
\newblock Gpt-4 technical report.
\newblock \emph{arXiv preprint arXiv:2303.08774}.

\bibitem[{Achiam et~al.(2017)Achiam, Held, Tamar, and Abbeel}]{achiam2017constrained}
Joshua Achiam, David Held, Aviv Tamar, and Pieter Abbeel. 2017.
\newblock Constrained policy optimization.
\newblock In \emph{International conference on machine learning}, pages 22--31. PMLR.

\bibitem[{Austin et~al.(2021)Austin, Odena, Nye, Bosma, Michalewski, Dohan, Jiang, Cai, Terry, Le et~al.}]{austin2021program}
Jacob Austin, Augustus Odena, Maxwell Nye, Maarten Bosma, Henryk Michalewski, David Dohan, Ellen Jiang, Carrie Cai, Michael Terry, Quoc Le, et~al. 2021.
\newblock Program synthesis with large language models.
\newblock \emph{arXiv preprint arXiv:2108.07732}.

\bibitem[{Banerjee et~al.(2024)Banerjee, Sahu, Luo, Vahldiek-Oberwagner, Yadwadkar, and Tiwari}]{banerjee2024sok}
Sarbartha Banerjee, Prateek Sahu, Mulong Luo, Anjo Vahldiek-Oberwagner, Neeraja~J Yadwadkar, and Mohit Tiwari. 2024.
\newblock Sok: A systems perspective on compound ai threats and countermeasures.
\newblock \emph{arXiv preprint arXiv:2411.13459}.

\bibitem[{Chase(2022)}]{Chase_LangChain_2022}
Harrison Chase. 2022.
\newblock \href {https://github.com/langchain-ai/langchain} {{LangChain}}.

\bibitem[{Chaudhry et~al.(2025)Chaudhry, Choukse, Goiri, Fonseca, Belay, and Bianchini}]{chaudhry2025towards}
Gohar~Irfan Chaudhry, Esha Choukse, {\'I}{\~n}igo Goiri, Rodrigo Fonseca, Adam Belay, and Ricardo Bianchini. 2025.
\newblock Towards resource-efficient compound ai systems.
\newblock \emph{arXiv preprint arXiv:2501.16634}.

\bibitem[{Chen et~al.(2025{\natexlab{a}})Chen, Davis, Hanin, Bailis, Zaharia, Zou, and Stoica}]{chen2025optimizing}
Lingjiao Chen, Jared~Quincy Davis, Boris Hanin, Peter Bailis, Matei Zaharia, James Zou, and Ion Stoica. 2025{\natexlab{a}}.
\newblock Optimizing model selection for compound ai systems.
\newblock \emph{arXiv preprint arXiv:2502.14815}.

\bibitem[{Chen et~al.(2021)Chen, Tworek, Jun, Yuan, Pinto, Kaplan, Edwards, Burda, Joseph, Brockman et~al.}]{chen2021evaluating}
Mark Chen, Jerry Tworek, Heewoo Jun, Qiming Yuan, Henrique Ponde De~Oliveira Pinto, Jared Kaplan, Harri Edwards, Yuri Burda, Nicholas Joseph, Greg Brockman, et~al. 2021.
\newblock Evaluating large language models trained on code.
\newblock \emph{arXiv preprint arXiv:2107.03374}.

\bibitem[{Chen et~al.(2025{\natexlab{b}})Chen, Jin, Deng, Chen, Huang, Yu, and Li}]{chen2025fedtextgrad}
Minghui Chen, Ruinan Jin, Wenlong Deng, Yuanyuan Chen, Zhi Huang, Han Yu, and Xiaoxiao Li. 2025{\natexlab{b}}.
\newblock \href {https://openreview.net/forum?id=Cy5IKvYbR3} {Can textual gradient work in federated learning?}
\newblock In \emph{The Thirteenth International Conference on Learning Representations}.

\bibitem[{Chen et~al.(2024)Chen, Foo, and Low}]{chen2024towards}
Zhiliang Chen, Chuan-Sheng Foo, and Bryan Kian~Hsiang Low. 2024.
\newblock Towards autoai: Optimizing a machine learning system with black-box and differentiable components.
\newblock In \emph{Forty-first International Conference on Machine Learning}.

\bibitem[{Cheng et~al.(2024)Cheng, Nie, and Swaminathan}]{cheng2024_trace}
Ching-An Cheng, Allen Nie, and Adith Swaminathan. 2024.
\newblock \href {https://openreview.net/forum?id=rYs2Dmn9tD} {Trace is the next autodiff: Generative optimization with rich feedback, execution traces, and {LLM}s}.
\newblock In \emph{The Thirty-eighth Annual Conference on Neural Information Processing Systems}.

\bibitem[{Cheridito et~al.(2022)Cheridito, Jentzen, Riekert, and Rossmannek}]{cheridito2022proof}
Patrick Cheridito, Arnulf Jentzen, Adrian Riekert, and Florian Rossmannek. 2022.
\newblock A proof of convergence for gradient descent in the training of artificial neural networks for constant target functions.
\newblock \emph{Journal of Complexity}, 72:101646.

\bibitem[{Cobbe et~al.(2021)Cobbe, Kosaraju, Bavarian, Chen, Jun, Kaiser, Plappert, Tworek, Hilton, Nakano, Hesse, and Schulman}]{cobbe2021trainingverifierssolvemath}
Karl Cobbe, Vineet Kosaraju, Mohammad Bavarian, Mark Chen, Heewoo Jun, Lukasz Kaiser, Matthias Plappert, Jerry Tworek, Jacob Hilton, Reiichiro Nakano, Christopher Hesse, and John Schulman. 2021.
\newblock \href {https://arxiv.org/abs/2110.14168} {Training verifiers to solve math word problems}.
\newblock \emph{Preprint}, arXiv:2110.14168.

\bibitem[{Dai et~al.(2023)Dai, Pan, Sun, Ji, Xu, Liu, Wang, and Yang}]{dai2023safe}
Josef Dai, Xuehai Pan, Ruiyang Sun, Jiaming Ji, Xinbo Xu, Mickel Liu, Yizhou Wang, and Yaodong Yang. 2023.
\newblock Safe rlhf: Safe reinforcement learning from human feedback.
\newblock \emph{arXiv preprint arXiv:2310.12773}.

\bibitem[{Das et~al.(2024)Das, Kamoi, Pang, Zhang, Xiong, and Zhang}]{das2024_greater}
Sarkar Snigdha~Sarathi Das, Ryo Kamoi, Bo~Pang, Yusen Zhang, Caiming Xiong, and Rui Zhang. 2024.
\newblock \href {https://arxiv.org/abs/2412.09722} {Greater: Gradients over reasoning makes smaller language models strong prompt optimizers}.
\newblock \emph{Preprint}, arXiv:2412.09722.

\bibitem[{Debenedetti et~al.(2024)Debenedetti, Severi, Carlini, Choquette-Choo, Jagielski, Nasr, Wallace, and Tram{\`e}r}]{debenedetti2024privacy}
Edoardo Debenedetti, Giorgio Severi, Nicholas Carlini, Christopher~A Choquette-Choo, Matthew Jagielski, Milad Nasr, Eric Wallace, and Florian Tram{\`e}r. 2024.
\newblock Privacy side channels in machine learning systems.
\newblock In \emph{33rd USENIX Security Symposium (USENIX Security 24)}, pages 6861--6848.

\bibitem[{Du et~al.(2023)Du, Li, Torralba, Tenenbaum, and Mordatch}]{du2023improving}
Yilun Du, Shuang Li, Antonio Torralba, Joshua~B Tenenbaum, and Igor Mordatch. 2023.
\newblock Improving factuality and reasoning in language models through multiagent debate.
\newblock In \emph{Forty-first International Conference on Machine Learning}.

\bibitem[{Gao et~al.(2025)Gao, Liu, He, Dou, Du, Deng, Hooi, Lin, and Pang}]{gao2025flowreasoner}
Hongcheng Gao, Yue Liu, Yufei He, Longxu Dou, Chao Du, Zhijie Deng, Bryan Hooi, Min Lin, and Tianyu Pang. 2025.
\newblock Flowreasoner: Reinforcing query-level meta-agents.
\newblock \emph{arXiv preprint arXiv:2504.15257}.

\bibitem[{Ghafarollahi and Buehler(2024)}]{ghafarollahi2024sciagents}
Alireza Ghafarollahi and Markus~J Buehler. 2024.
\newblock Sciagents: Automating scientific discovery through multi-agent intelligent graph reasoning.
\newblock \emph{arXiv preprint arXiv:2409.05556}.

\bibitem[{Guo et~al.(2025)Guo, Wang, Guo, Li, Song, Tan, Liu, Bian, and Yang}]{EvoPropmt}
Qingyan Guo, Rui Wang, Junliang Guo, Bei Li, Kaitao Song, Xu~Tan, Guoqing Liu, Jiang Bian, and Yujiu Yang. 2025.
\newblock \href {https://arxiv.org/abs/2309.08532} {Evoprompt: Connecting llms with evolutionary algorithms yields powerful prompt optimizers}.
\newblock \emph{Preprint}, arXiv:2309.08532.

\bibitem[{Han et~al.(2024)Han, Hu, Shah, Jin, Yao, Stripelis, Xu, and He}]{han2024torchopera}
Shanshan Han, Zijian Hu, Alay~Dilipbhai Shah, Han Jin, Yuhang Yao, Dimitris Stripelis, Zhaozhuo Xu, and Chaoyang He. 2024.
\newblock Torchopera: A compound ai system for llm safety.
\newblock \emph{arXiv preprint arXiv:2406.10847}.

\bibitem[{Hendrycks et~al.(2021)Hendrycks, Burns, Basart, Zou, Mazeika, Song, and Steinhardt}]{hendrycks2021measuringmassivemultitasklanguage}
Dan Hendrycks, Collin Burns, Steven Basart, Andy Zou, Mantas Mazeika, Dawn Song, and Jacob Steinhardt. 2021.
\newblock \href {https://arxiv.org/abs/2009.03300} {Measuring massive multitask language understanding}.
\newblock \emph{Preprint}, arXiv:2009.03300.

\bibitem[{Hong et~al.(2023)Hong, Zheng, Chen, Cheng, Wang, Zhang, Wang, Yau, Lin, Zhou et~al.}]{hong2023metagpt}
Sirui Hong, Xiawu Zheng, Jonathan Chen, Yuheng Cheng, Jinlin Wang, Ceyao Zhang, Zili Wang, Steven Ka~Shing Yau, Zijuan Lin, Liyang Zhou, et~al. 2023.
\newblock Metagpt: Meta programming for multi-agent collaborative framework.
\newblock \emph{arXiv preprint arXiv:2308.00352}, 3(4):6.

\bibitem[{Hu et~al.(2024)Hu, Lu, and Clune}]{hu2024_ADAS}
Shengran Hu, Cong Lu, and Jeff Clune. 2024.
\newblock Automated design of agentic systems.
\newblock \emph{arXiv preprint arXiv:2408.08435}.

\bibitem[{Hutzenthaler et~al.(2021)Hutzenthaler, Jentzen, Pohl, Riekert, and Scarpa}]{hutzenthaler2021convergence}
Martin Hutzenthaler, Arnulf Jentzen, Katharina Pohl, Adrian Riekert, and Luca Scarpa. 2021.
\newblock Convergence proof for stochastic gradient descent in the training of deep neural networks with relu activation for constant target functions.
\newblock \emph{arXiv preprint arXiv:2112.07369}.

\bibitem[{Khattab et~al.(2023)Khattab, Singhvi, Maheshwari, Zhang, Santhanam, Vardhamanan, Haq, Sharma, Joshi, Moazam, Miller, Zaharia, and Potts}]{khattab2023_dspy}
Omar Khattab, Arnav Singhvi, Paridhi Maheshwari, Zhiyuan Zhang, Keshav Santhanam, Sri Vardhamanan, Saiful Haq, Ashutosh Sharma, Thomas~T. Joshi, Hanna Moazam, Heather Miller, Matei Zaharia, and Christopher Potts. 2023.
\newblock \href {https://arxiv.org/abs/2310.03714} {Dspy: Compiling declarative language model calls into self-improving pipelines}.
\newblock \emph{Preprint}, arXiv:2310.03714.

\bibitem[{Li et~al.(2025)Li, Hu, Qu, Li, and Cheng}]{li2025TPO}
Yafu Li, Xuyang Hu, Xiaoye Qu, Linjie Li, and Yu~Cheng. 2025.
\newblock \href {https://arxiv.org/abs/2501.12895} {Test-time preference optimization: On-the-fly alignment via iterative textual feedback}.
\newblock \emph{Preprint}, arXiv:2501.12895.

\bibitem[{Li et~al.(2022)Li, Choi, Chung, Kushman, Schrittwieser, Leblond, Eccles, Keeling, Gimeno, Dal~Lago et~al.}]{li2022competition}
Yujia Li, David Choi, Junyoung Chung, Nate Kushman, Julian Schrittwieser, R{\'e}mi Leblond, Tom Eccles, James Keeling, Felix Gimeno, Agustin Dal~Lago, et~al. 2022.
\newblock Competition-level code generation with alphacode.
\newblock \emph{Science}, 378(6624):1092--1097.

\bibitem[{Li et~al.(2024)Li, Xu, Mei, Hua, Rama, Raheja, Wang, Zhu, and Zhang}]{li2024autoflow}
Zelong Li, Shuyuan Xu, Kai Mei, Wenyue Hua, Balaji Rama, Om~Raheja, Hao Wang, He~Zhu, and Yongfeng Zhang. 2024.
\newblock Autoflow: Automated workflow generation for large language model agents.
\newblock \emph{arXiv preprint arXiv:2407.12821}.

\bibitem[{Lin et~al.(2024)Lin, Sheng, Zhao, Wang, Yue, Wu, Liu, Liu, Huang, and Liu}]{lin2024_llm_opt}
Matthieu Lin, Jenny Sheng, Andrew Zhao, Shenzhi Wang, Yang Yue, Yiran Wu, Huan Liu, Jun Liu, Gao Huang, and Yong-Jin Liu. 2024.
\newblock Llm-based optimization of compound ai systems: A survey.
\newblock \emph{arXiv preprint arXiv:2410.16392}.

\bibitem[{Liu et~al.(2025)Liu, Li, Zhang, Wang, He, Hong, Liu, Zhang, Song, Zhu, Cheng, Wang, Wang, Luo, Jin, Zhang, Liu, Chen, Zhang, Yu, Shi, Li, Wu, Teng, Jia, Xu, Xiang, Lin, Liu, Liu, Su, Sun, Berseth, Nie, Foster, Ward, Wu, Gu, Zhuge, Tang, Wang, You, Wang, Pei, Yang, Qi, and Wu}]{liu2025advancesagents}
Bang Liu, Xinfeng Li, Jiayi Zhang, Jinlin Wang, Tanjin He, Sirui Hong, Hongzhang Liu, Shaokun Zhang, Kaitao Song, Kunlun Zhu, Yuheng Cheng, Suyuchen Wang, Xiaoqiang Wang, Yuyu Luo, Haibo Jin, Peiyan Zhang, Ollie Liu, Jiaqi Chen, Huan Zhang, Zhaoyang Yu, Haochen Shi, Boyan Li, Dekun Wu, Fengwei Teng, Xiaojun Jia, Jiawei Xu, Jinyu Xiang, Yizhang Lin, Tianming Liu, Tongliang Liu, Yu~Su, Huan Sun, Glen Berseth, Jianyun Nie, Ian Foster, Logan Ward, Qingyun Wu, Yu~Gu, Mingchen Zhuge, Xiangru Tang, Haohan Wang, Jiaxuan You, Chi Wang, Jian Pei, Qiang Yang, Xiaoliang Qi, and Chenglin Wu. 2025.
\newblock \href {https://arxiv.org/abs/2504.01990} {Advances and challenges in foundation agents: From brain-inspired intelligence to evolutionary, collaborative, and safe systems}.
\newblock \emph{Preprint}, arXiv:2504.01990.

\bibitem[{Liu(2022)}]{Liu_LlamaIndex_2022}
Jerry Liu. 2022.
\newblock \href {https://doi.org/10.5281/zenodo.1234} {{LlamaIndex}}.

\bibitem[{Liu et~al.(2023)Liu, Yu, Zhang, Xu, Lei, Lai, Gu, Ding, Men, Yang, Zhang, Deng, Zeng, Du, Zhang, Shen, Zhang, Su, Sun, Huang, Dong, and Tang}]{liu2023agentbench}
Xiao Liu, Hao Yu, Hanchen Zhang, Yifan Xu, Xuanyu Lei, Hanyu Lai, Yu~Gu, Hangliang Ding, Kaiwen Men, Kejuan Yang, Shudan Zhang, Xiang Deng, Aohan Zeng, Zhengxiao Du, Chenhui Zhang, Sheng Shen, Tianjun Zhang, Yu~Su, Huan Sun, Minlie Huang, Yuxiao Dong, and Jie Tang. 2023.
\newblock Agentbench: Evaluating llms as agents.
\newblock \emph{arXiv preprint arXiv: 2308.03688}.

\bibitem[{Liu et~al.(2024)Liu, Zhang, Li, Liu, and Yang}]{DyLAN}
Zijun Liu, Yanzhe Zhang, Peng Li, Yang Liu, and Diyi Yang. 2024.
\newblock A dynamic llm-powered agent network for task-oriented agent collaboration.
\newblock In \emph{First Conference on Language Modeling}.

\bibitem[{Mialon et~al.(2023)Mialon, Fourrier, Wolf, LeCun, and Scialom}]{mialon2023gaia}
Gr{\'e}goire Mialon, Cl{\'e}mentine Fourrier, Thomas Wolf, Yann LeCun, and Thomas Scialom. 2023.
\newblock Gaia: a benchmark for general ai assistants.
\newblock In \emph{The Twelfth International Conference on Learning Representations}.

\bibitem[{Nie et~al.(2025)Nie, Feng, Ye, Liang, Lu, Yao, Alahi, and Zou}]{nie2025weak}
Fan Nie, Lan Feng, Haotian Ye, Weixin Liang, Pan Lu, Huaxiu Yao, Alexandre Alahi, and James Zou. 2025.
\newblock Weak-for-strong: Training weak meta-agent to harness strong executors.
\newblock \emph{arXiv preprint arXiv:2504.04785}.

\bibitem[{Novac et~al.(2022)Novac, Chirodea, Novac, Bizon, Oproescu, Stan, and Gordan}]{novac2022analysis}
Ovidiu-Constantin Novac, Mihai~Cristian Chirodea, Cornelia~Mihaela Novac, Nicu Bizon, Mihai Oproescu, Ovidiu~Petru Stan, and Cornelia~Emilia Gordan. 2022.
\newblock Analysis of the application efficiency of tensorflow and pytorch in convolutional neural network.
\newblock \emph{Sensors}, 22(22):8872.

\bibitem[{Opsahl-Ong et~al.(2024)Opsahl-Ong, Ryan, Purtell, Broman, Potts, Zaharia, and Khattab}]{opsahl2024optimizing}
Krista Opsahl-Ong, Michael~J Ryan, Josh Purtell, David Broman, Christopher Potts, Matei Zaharia, and Omar Khattab. 2024.
\newblock Optimizing instructions and demonstrations for multi-stage language model programs.
\newblock \emph{arXiv preprint arXiv:2406.11695}.

\bibitem[{Park et~al.(2025)Park, Han, Guo, Ozdaglar, Zhang, and Kim}]{park2025_maporl}
Chanwoo Park, Seungju Han, Xingzhi Guo, Asuman Ozdaglar, Kaiqing Zhang, and Joo-Kyung Kim. 2025.
\newblock \href {https://arxiv.org/abs/2502.18439} {Maporl: Multi-agent post-co-training for collaborative large language models with reinforcement learning}.
\newblock \emph{Preprint}, arXiv:2502.18439.

\bibitem[{Paszke(2019)}]{paszke2019pytorch}
A~Paszke. 2019.
\newblock Pytorch: An imperative style, high-performance deep learning library.
\newblock \emph{arXiv preprint arXiv:1912.01703}.

\bibitem[{Patel et~al.(2024)Patel, Chakraborty, Suttle, Wang, Bedi, and Manocha}]{patel2024_aime}
Bhrij Patel, Souradip Chakraborty, Wesley~A. Suttle, Mengdi Wang, Amrit~Singh Bedi, and Dinesh Manocha. 2024.
\newblock \href {https://arxiv.org/abs/2410.03131} {Aime: Ai system optimization via multiple llm evaluators}.
\newblock \emph{Preprint}, arXiv:2410.03131.

\bibitem[{Pryzant et~al.(2023)Pryzant, Iter, Li, Lee, Zhu, and Zeng}]{APO}
Reid Pryzant, Dan Iter, Jerry Li, Yin~Tat Lee, Chenguang Zhu, and Michael Zeng. 2023.
\newblock Automatic prompt optimization with" gradient descent" and beam search.
\newblock \emph{arXiv preprint arXiv:2305.03495}.

\bibitem[{Rafailov et~al.(2023)Rafailov, Sharma, Mitchell, Manning, Ermon, and Finn}]{DPO}
Rafael Rafailov, Archit Sharma, Eric Mitchell, Christopher~D Manning, Stefano Ermon, and Chelsea Finn. 2023.
\newblock Direct preference optimization: Your language model is secretly a reward model.
\newblock \emph{Advances in Neural Information Processing Systems}, 36:53728--53741.

\bibitem[{Rombach et~al.(2022)Rombach, Blattmann, Lorenz, Esser, and Ommer}]{rombach2022high}
Robin Rombach, Andreas Blattmann, Dominik Lorenz, Patrick Esser, and Bj{\"o}rn Ommer. 2022.
\newblock High-resolution image synthesis with latent diffusion models.
\newblock In \emph{Proceedings of the IEEE/CVF conference on computer vision and pattern recognition}, pages 10684--10695.

\bibitem[{Rumelhart et~al.(1986)Rumelhart, Hinton, and Williams}]{rumelhart1986learning}
David~E Rumelhart, Geoffrey~E Hinton, and Ronald~J Williams. 1986.
\newblock Learning representations by back-propagating errors.
\newblock \emph{nature}, 323(6088):533--536.

\bibitem[{Santhanam et~al.(2024)Santhanam, Raghavan, Rahman, Venkatesh, Kunjal, Thaker, Levis, and Zaharia}]{santhanam2024alto}
Keshav Santhanam, Deepti Raghavan, Muhammad~Shahir Rahman, Thejas Venkatesh, Neha Kunjal, Pratiksha Thaker, Philip Levis, and Matei Zaharia. 2024.
\newblock Alto: An efficient network orchestrator for compound ai systems.
\newblock In \emph{Proceedings of the 4th Workshop on Machine Learning and Systems}, pages 117--125.

\bibitem[{Sordoni et~al.(2023)Sordoni, Yuan, C{\^o}t{\'e}, Pereira, Trischler, Xiao, Hosseini, Niedtner, and Le~Roux}]{sordoni2023joint}
Alessandro Sordoni, Eric Yuan, Marc-Alexandre C{\^o}t{\'e}, Matheus Pereira, Adam Trischler, Ziang Xiao, Arian Hosseini, Friederike Niedtner, and Nicolas Le~Roux. 2023.
\newblock Joint prompt optimization of stacked llms using variational inference.
\newblock \emph{Advances in Neural Information Processing Systems}, 36:58128--58151.

\bibitem[{Soylu et~al.(2024)Soylu, Potts, and Khattab}]{soylu2024fine}
Dilara Soylu, Christopher Potts, and Omar Khattab. 2024.
\newblock Fine-tuning and prompt optimization: Two great steps that work better together.
\newblock \emph{arXiv preprint arXiv:2407.10930}.

\bibitem[{Su et~al.(2025)Su, Xia, Shi, Wang, Huang, Wang, Shi, Jingsong, and He}]{su2025debflowautomatingagentcreation}
Jinwei Su, Yinghui Xia, Ronghua Shi, Jianhui Wang, Jianuo Huang, Yijin Wang, Tianyu Shi, Yang Jingsong, and Lewei He. 2025.
\newblock \href {https://arxiv.org/abs/2503.23781} {Debflow: Automating agent creation via agent debate}.
\newblock \emph{Preprint}, arXiv:2503.23781.

\bibitem[{Subramaniam et~al.(2025)Subramaniam, Du, Tenenbaum, Torralba, Li, and Mordatch}]{subramaniam2025multiagentFT}
Vighnesh Subramaniam, Yilun Du, Joshua~B. Tenenbaum, Antonio Torralba, Shuang Li, and Igor Mordatch. 2025.
\newblock \href {https://openreview.net/forum?id=JtGPIZpOrz} {Multiagent finetuning: Self improvement with diverse reasoning chains}.
\newblock In \emph{The Thirteenth International Conference on Learning Representations}.

\bibitem[{Trinh et~al.(2024)Trinh, Wu, Le, He, and Luong}]{trinh2024solving}
Trieu~H Trinh, Yuhuai Wu, Quoc~V Le, He~He, and Thang Luong. 2024.
\newblock Solving olympiad geometry without human demonstrations.
\newblock \emph{Nature}, 625(7995):476--482.

\bibitem[{Trirat et~al.(2024)Trirat, Jeong, and Hwang}]{trirat2024autoML}
Patara Trirat, Wonyong Jeong, and Sung~Ju Hwang. 2024.
\newblock \href {https://arxiv.org/abs/2410.02958} {Automl-agent: A multi-agent llm framework for full-pipeline automl}.
\newblock \emph{Preprint}, arXiv:2410.02958.

\bibitem[{Vaswani et~al.(2017)Vaswani, Shazeer, Parmar, Uszkoreit, Jones, Gomez, Kaiser, and Polosukhin}]{vaswani2017attention}
Ashish Vaswani, Noam Shazeer, Niki Parmar, Jakob Uszkoreit, Llion Jones, Aidan~N Gomez, {\L}ukasz Kaiser, and Illia Polosukhin. 2017.
\newblock Attention is all you need.
\newblock \emph{Advances in neural information processing systems}, 30.

\bibitem[{Wang et~al.(2024{\natexlab{a}})Wang, Alyahya, Ashley, Serikov, Khizbullin, Faccio, and Schmidhuber}]{wang2024correctly}
Wenyi Wang, Hisham~A Alyahya, Dylan~R Ashley, Oleg Serikov, Dmitrii Khizbullin, Francesco Faccio, and J{\"u}rgen Schmidhuber. 2024{\natexlab{a}}.
\newblock How to correctly do semantic backpropagation on language-based agentic systems.
\newblock \emph{arXiv preprint arXiv:2412.03624}.

\bibitem[{Wang et~al.(2024{\natexlab{b}})Wang, Alyahya, Ashley, Serikov, Khizbullin, Faccio, and Schmidhuber}]{wang2024semantic_backpropagation}
Wenyi Wang, Hisham~A. Alyahya, Dylan~R. Ashley, Oleg Serikov, Dmitrii Khizbullin, Francesco Faccio, and Jürgen Schmidhuber. 2024{\natexlab{b}}.
\newblock \href {https://arxiv.org/abs/2412.03624} {How to correctly do semantic backpropagation on language-based agentic systems}.
\newblock \emph{Preprint}, arXiv:2412.03624.

\bibitem[{Wang et~al.(2025{\natexlab{a}})Wang, Zhang, Ding, Tsai, and Koyejo}]{wang2025_systemDPO}
Xiangwen Wang, Yibo~Jacky Zhang, Zhoujie Ding, Katherine Tsai, and Sanmi Koyejo. 2025{\natexlab{a}}.
\newblock Aligning compound ai systems via system-level dpo.
\newblock \emph{arXiv preprint arXiv:2502.17721}.

\bibitem[{Wang et~al.(2025{\natexlab{b}})Wang, Yang, Li, Wang, and Aragam}]{wang2025_scoreflow}
Yinjie Wang, Ling Yang, Guohao Li, Mengdi Wang, and Bryon Aragam. 2025{\natexlab{b}}.
\newblock Scoreflow: Mastering llm agent workflows via score-based preference optimization.
\newblock \emph{arXiv preprint arXiv:2502.04306}.

\bibitem[{Wei et~al.(2025)Wei, Huang, Li, Xing, Lin, and He}]{wei2025vflow}
Yangbo Wei, Zhen Huang, Huang Li, Wei~W. Xing, Ting-Jung Lin, and Lei He. 2025.
\newblock \href {https://arxiv.org/abs/2504.03723} {Vflow: Discovering optimal agentic workflows for verilog generation}.
\newblock \emph{Preprint}, arXiv:2504.03723.

\bibitem[{Williams(1992)}]{williams1992simple}
Ronald~J Williams. 1992.
\newblock Simple statistical gradient-following algorithms for connectionist reinforcement learning.
\newblock \emph{Machine learning}, 8:229--256.

\bibitem[{Xi et~al.(2024)Xi, Ding, Chen, Hong, Guo, Wang, Yang, Liao, Guo, He, Gao, Chen, Zheng, Zou, Gui, Zhang, Qiu, Huang, Wu, and Jiang}]{xi2024agentgym}
Zhiheng Xi, Yiwen Ding, Wenxiang Chen, Boyang Hong, Honglin Guo, Junzhe Wang, Dingwen Yang, Chenyang Liao, Xin Guo, Wei He, Songyang Gao, Lu~Chen, Rui Zheng, Yicheng Zou, Tao Gui, Qi~Zhang, Xipeng Qiu, Xuanjing Huang, Zuxuan Wu, and Yu-Gang Jiang. 2024.
\newblock \href {https://arxiv.org/abs/2406.04151} {Agentgym: Evolving large language model-based agents across diverse environments}.
\newblock \emph{Preprint}, arXiv:2406.04151.

\bibitem[{Xia et~al.(2024)Xia, Deng, Dunn, and Zhang}]{xia2024agentless}
Chunqiu~Steven Xia, Yinlin Deng, Soren Dunn, and Lingming Zhang. 2024.
\newblock Agentless: Demystifying llm-based software engineering agents.
\newblock \emph{arXiv preprint arXiv:2407.01489}.

\bibitem[{Xu et~al.(2024{\natexlab{a}})Xu, Yuksekgonul, Guestrin, and Zou}]{xu2024metatextgrad}
Guowei Xu, Mert Yuksekgonul, Carlos Guestrin, and James Zou. 2024{\natexlab{a}}.
\newblock \href {https://openreview.net/forum?id=yzieYIT9hu} {metatextgrad: Learning to learn with language models as optimizers}.
\newblock In \emph{Adaptive Foundation Models: Evolving AI for Personalized and Efficient Learning}.

\bibitem[{Xu et~al.(2024{\natexlab{b}})Xu, Li, Mei, and Zhang}]{xu2024core}
Shuyuan Xu, Zelong Li, Kai Mei, and Yongfeng Zhang. 2024{\natexlab{b}}.
\newblock Core: Llm as interpreter for natural language programming, pseudo-code programming, and flow programming of ai agents.
\newblock \emph{arXiv e-prints}, pages arXiv--2405.

\bibitem[{Yang et~al.(2024{\natexlab{a}})Yang, Lu, Wang, and Zhou}]{yang2024ai45circlawroadmaptrustworthy}
Chao Yang, Chaochao Lu, Yingchun Wang, and Bowen Zhou. 2024{\natexlab{a}}.
\newblock \href {https://arxiv.org/abs/2412.14186} {Towards ai-$45^{\circ}$ law: A roadmap to trustworthy agi}.
\newblock \emph{Preprint}, arXiv:2412.14186.

\bibitem[{Yang et~al.(2023)Yang, Wang, Lu, Liu, Le, Zhou, and Chen}]{OPRO}
Chengrun Yang, Xuezhi Wang, Yifeng Lu, Hanxiao Liu, Quoc~V Le, Denny Zhou, and Xinyun Chen. 2023.
\newblock Large language models as optimizers.
\newblock \emph{arXiv preprint arXiv:2309.03409}.

\bibitem[{Yang et~al.(2024{\natexlab{b}})Yang, Jimenez, Wettig, Lieret, Yao, Narasimhan, and Press}]{yang2024swe}
John Yang, Carlos Jimenez, Alexander Wettig, Kilian Lieret, Shunyu Yao, Karthik Narasimhan, and Ofir Press. 2024{\natexlab{b}}.
\newblock Swe-agent: Agent-computer interfaces enable automated software engineering.
\newblock \emph{Advances in Neural Information Processing Systems}, 37:50528--50652.

\bibitem[{Ye et~al.(2025)Ye, Tang, Ge, Du, Yin, Chen, and Shao}]{ye2025_masgpttrainingllmsbuild}
Rui Ye, Shuo Tang, Rui Ge, Yaxin Du, Zhenfei Yin, Siheng Chen, and Jing Shao. 2025.
\newblock Mas-gpt: Training llms to build llm-based multi-agent systems.
\newblock \emph{arXiv preprint arXiv:2503.03686}.

\bibitem[{Yi et~al.(2024)Yi, Liu, Sun, Cong, He, Song, Xu, and Li}]{yi2024jailbreak}
Sibo Yi, Yule Liu, Zhen Sun, Tianshuo Cong, Xinlei He, Jiaxing Song, Ke~Xu, and Qi~Li. 2024.
\newblock Jailbreak attacks and defenses against large language models: A survey.
\newblock \emph{arXiv preprint arXiv:2407.04295}.

\bibitem[{Yin and Wang(2025)}]{yin2025_llm_autodiff}
Li~Yin and Zhangyang Wang. 2025.
\newblock \href {https://arxiv.org/abs/2501.16673} {Llm-autodiff: Auto-differentiate any llm workflow}.
\newblock \emph{Preprint}, arXiv:2501.16673.

\bibitem[{Yuksekgonul et~al.(2025)Yuksekgonul, Bianchi, Boen, Liu, Lu, Huang, Guestrin, and Zou}]{Mert2024_textgrad}
Mert Yuksekgonul, Federico Bianchi, Joseph Boen, Sheng Liu, Pan Lu, Zhi Huang, Carlos Guestrin, and James Zou. 2025.
\newblock Optimizing generative ai by backpropagating language model feedback.
\newblock \emph{Nature}, 639(8055):609--616.

\bibitem[{Zaharia et~al.(2024)Zaharia, Khattab, Chen, Davis, Miller, Potts, Zou, Carbin, Frankle, Rao, and Ghodsi}]{compound-ai-blog}
Matei Zaharia, Omar Khattab, Lingjiao Chen, Jared~Quincy Davis, Heather Miller, Chris Potts, James Zou, Michael Carbin, Jonathan Frankle, Naveen Rao, and Ali Ghodsi. 2024.
\newblock The shift from models to compound ai systems.
\newblock \url{https://bair.berkeley.edu/blog/2024/02/18/compound-ai-systems/}.

\bibitem[{Zhang et~al.(2025)Zhang, Niu, Fang, Wang, Bai, and Wang}]{zhang2025multi}
Guibin Zhang, Luyang Niu, Junfeng Fang, Kun Wang, Lei Bai, and Xiang Wang. 2025.
\newblock Multi-agent architecture search via agentic supernet.
\newblock \emph{arXiv preprint arXiv:2502.04180}.

\bibitem[{Zhang et~al.(2024{\natexlab{a}})Zhang, Xiang, Yu, Teng, Chen, Chen, Zhuge, Cheng, Hong, Wang et~al.}]{zhang2024_aflow}
Jiayi Zhang, Jinyu Xiang, Zhaoyang Yu, Fengwei Teng, Xionghui Chen, Jiaqi Chen, Mingchen Zhuge, Xin Cheng, Sirui Hong, Jinlin Wang, et~al. 2024{\natexlab{a}}.
\newblock Aflow: Automating agentic workflow generation.
\newblock \emph{arXiv preprint arXiv:2410.10762}.

\bibitem[{Zhang et~al.(2024{\natexlab{b}})Zhang, Jin, Hu, Li, Kang, Luo, Song, and Wang}]{zhang2024_revolve}
Peiyan Zhang, Haibo Jin, Leyang Hu, Xinnuo Li, Liying Kang, Man Luo, Yangqiu Song, and Haohan Wang. 2024{\natexlab{b}}.
\newblock \href {https://arxiv.org/abs/2412.03092} {Revolve: Optimizing ai systems by tracking response evolution in textual optimization}.
\newblock \emph{Preprint}, arXiv:2412.03092.

\bibitem[{Zhang et~al.(2024{\natexlab{c}})Zhang, Sun, Chen, Pfister, Zhang, and Arik}]{zhang2024chain}
Yusen Zhang, Ruoxi Sun, Yanfei Chen, Tomas Pfister, Rui Zhang, and Sercan Arik. 2024{\natexlab{c}}.
\newblock Chain of agents: Large language models collaborating on long-context tasks.
\newblock \emph{Advances in Neural Information Processing Systems}, 37:132208--132237.

\bibitem[{Zhao et~al.(2025)Zhao, Yuksekgonul, Wu, and Zou}]{zhao2025_sirius}
Wanjia Zhao, Mert Yuksekgonul, Shirley Wu, and James Zou. 2025.
\newblock \href {https://arxiv.org/abs/2502.04780} {Sirius: Self-improving multi-agent systems via bootstrapped reasoning}.
\newblock \emph{Preprint}, arXiv:2502.04780.

\bibitem[{Zheng et~al.(2025)Zheng, Chen, Lyu, Ng, Zhang, Ong, Tsang, and Yin}]{zheng2025mermaidflow}
Chengqi Zheng, Jianda Chen, Yueming Lyu, Wen Zheng~Terence Ng, Haopeng Zhang, Yew-Soon Ong, Ivor Tsang, and Haiyan Yin. 2025.
\newblock Mermaidflow: Redefining agentic workflow generation via safety-constrained evolutionary programming.
\newblock \emph{arXiv preprint arXiv:2505.22967}.

\bibitem[{Zhou et~al.(2022{\natexlab{a}})Zhou, Sch{\"a}rli, Hou, Wei, Scales, Wang, Schuurmans, Cui, Bousquet, Le et~al.}]{zhou2022least}
Denny Zhou, Nathanael Sch{\"a}rli, Le~Hou, Jason Wei, Nathan Scales, Xuezhi Wang, Dale Schuurmans, Claire Cui, Olivier Bousquet, Quoc Le, et~al. 2022{\natexlab{a}}.
\newblock Least-to-most prompting enables complex reasoning in large language models.
\newblock \emph{arXiv preprint arXiv:2205.10625}.

\bibitem[{Zhou et~al.(2025)Zhou, Wan, Sun, Palangi, Iqbal, Vuli{\'c}, Korhonen, and Ar{\i}k}]{zhou2025_multi}
Han Zhou, Xingchen Wan, Ruoxi Sun, Hamid Palangi, Shariq Iqbal, Ivan Vuli{\'c}, Anna Korhonen, and Sercan~{\"O} Ar{\i}k. 2025.
\newblock Multi-agent design: Optimizing agents with better prompts and topologies.
\newblock \emph{arXiv preprint arXiv:2502.02533}.

\bibitem[{Zhou et~al.(2024)Zhou, Ou, Ding, Li, Wu, Wang, Chen, Wang, Xu, Zhang et~al.}]{zhou2024_symbolic}
Wangchunshu Zhou, Yixin Ou, Shengwei Ding, Long Li, Jialong Wu, Tiannan Wang, Jiamin Chen, Shuai Wang, Xiaohua Xu, Ningyu Zhang, et~al. 2024.
\newblock Symbolic learning enables self-evolving agents.
\newblock \emph{arXiv preprint arXiv:2406.18532}.

\bibitem[{Zhou et~al.(2022{\natexlab{b}})Zhou, Muresanu, Han, Paster, Pitis, Chan, and Ba}]{APE}
Yongchao Zhou, Andrei~Ioan Muresanu, Ziwen Han, Keiran Paster, Silviu Pitis, Harris Chan, and Jimmy Ba. 2022{\natexlab{b}}.
\newblock Large language models are human-level prompt engineers.
\newblock In \emph{The Eleventh International Conference on Learning Representations}.

\bibitem[{Zhuge et~al.(2023)Zhuge, Liu, Faccio, Ashley, Csord{\'a}s, Gopalakrishnan, Hamdi, Hammoud, Herrmann, Irie et~al.}]{zhuge2023mindstorms}
Mingchen Zhuge, Haozhe Liu, Francesco Faccio, Dylan~R Ashley, R{\'o}bert Csord{\'a}s, Anand Gopalakrishnan, Abdullah Hamdi, Hasan Abed Al~Kader Hammoud, Vincent Herrmann, Kazuki Irie, et~al. 2023.
\newblock Mindstorms in natural language-based societies of mind.
\newblock \emph{arXiv preprint arXiv:2305.17066}.

\bibitem[{Zhuge et~al.(2024)Zhuge, Wang, Kirsch, Faccio, Khizbullin, and Schmidhuber}]{zhuge2024_gptswarm}
Mingchen Zhuge, Wenyi Wang, Louis Kirsch, Francesco Faccio, Dmitrii Khizbullin, and J{\"u}rgen Schmidhuber. 2024.
\newblock Gptswarm: Language agents as optimizable graphs.
\newblock In \emph{Forty-first International Conference on Machine Learning}.

\end{thebibliography}
